%% file: 0-crt-neurips-camera-ready.tex
\documentclass{article}

\usepackage[final]{neurips_2021}
\usepackage[utf8]{inputenc} % allow utf-8 input
\usepackage[T1]{fontenc}    % use 8-bit T1 fonts
\usepackage{hyperref}       % hyperlinks
\usepackage{url}            % simple URL typesetting
\usepackage{booktabs}       % professional-quality tables
\usepackage{amsfonts}       % blackboard math symbols
\usepackage{nicefrac}       % compact symbols for 1/2, etc.
\usepackage{microtype}      % microtypography
\usepackage{xcolor}         % colors
\usepackage{amsmath}
\usepackage{amssymb}
\usepackage{algorithm,algorithmic}
\usepackage{graphicx}

\usepackage{natbib}
\setcitestyle{square,sort,comma,numbers}

\usepackage{amsthm,amscd,amstext}

\usepackage{wrapfig}
\usepackage{booktabs}       % professional-quality tables

\usepackage{multirow}

\usepackage{amsmath}

\include{math_cmds}

\newcommand{\GCL}{\text{GCL}}

\newcommand{\FLOW}{\text{FLOW}}

\newcommand{\Sv}{\bs{S}}
\newcommand{\Zv}{\bs{Z}}

\newcommand{\CO}{\mathcal{O}}

\title{Supercharging Imbalanced Data Learning With \\ Energy-based Contrastive Representation Transfer}

\author{%
  Zidi Xiu$\thanks{Contributed equally.}$ \And
  Junya Chen${}^{*}$ \And
  Benjamin Goldstein \And
  Ricardo Henao \And
  Lawrence Carin \And
  Chenyang Tao${}^{\dagger}$ \\
  Duke University\\
  \texttt{\{zidi.xiu, junya.chen, chenyang.tao\}@duke.edu} \\
}

\begin{document}

\maketitle

\begin{abstract}
Dealing with severe class imbalance poses a major challenge for many real-world applications, especially when the accurate classification and generalization of minority classes are of primary interest.
In computer vision and NLP, learning from datasets with long-tail behavior is a recurring theme, especially for naturally occurring labels.
Existing solutions mostly appeal to sampling or weighting adjustments to alleviate the extreme imbalance, or impose inductive bias to prioritize generalizable associations.
Here we take a novel perspective to promote sample efficiency and model generalization based on the invariance principles of causality.
Our contribution posits a meta-distributional scenario, where the causal generating mechanism for label-conditional features is invariant across different labels.
Such causal assumption enables efficient knowledge transfer from the dominant classes to their under-represented counterparts, even if their feature distributions show apparent disparities.
This allows us to leverage a causal data augmentation procedure to enlarge the representation of minority classes.
Our development is orthogonal to the existing imbalanced data learning techniques thus can be seamlessly integrated.
The proposed approach is validated on an extensive set of synthetic and real-world tasks against state-of-the-art solutions. 
% ({\it e.g.}, financial fraud, severe weather, rare disease, traffic safety, etc.)
\end{abstract}

\vspace{-8pt}
\section{Introduction}
\vspace{-4pt}
\noindent Learning with imbalanced datasets is a common yet still very challenging scenario in many machine learning applications.
Typical scenarios include: ($i$) rare events, where the event prevalence is extremely low while their implications are of high cost, {\it e.g.}, severe risks that people seek to avert \cite{king2001logistic}; ($ii$) emerging objects in a dynamic environment, which call for quick adaptation of an agent to identify new cases with only a handful target examples and plentiful past experience \cite{fei2006one}.
A typical scenario in natural datasets is that the occurrence of different objects follows a power law distribution. 
And in many situations, the accurate identification of those rarer instances bears more significant social-economic values, {\it e.g.}, fraud detection \cite{dal2017credit}, driving safety \cite{dingus2016driver}, nature conservation \cite{kaggle2017planet}, social fairness \cite{chouldechova2018frontiers}, and public health \cite{machado2020rare, xiu2020survival}. 

%\textcolor{red}{Concrete examples bearing significant social-economic values include vehicle collisions in traffic safety \cite{dingus2016driver}, fraudulent activities in financial business \cite{dal2017credit}, minority fairness in social justice \cite{chouldechova2018frontiers} and emerging crisis in public health \cite{machado2020rare}. }

Notably, severe class imbalance and lack of minority labels are the two major difficulties in this setting, which render standard learning strategies unsuitable \cite{kotsiantis2006handling}.  
Without explicit statistical adjustments, the imbalance induces bias towards the majority classes.
On the other hand, the lack of minority representations prevents the identification of stable correlations that generalize in predictive settings.
In addition, due to technological advancements, more and higher dimensional data are routinely collected for analysis.
This also inadvertently exacerbates the issue of minority modeling as there is an excess of predictors relative to the limited occurrence of minority samples.

\begin{figure}[t!]
\begin{center}
\includegraphics[width=\textwidth]{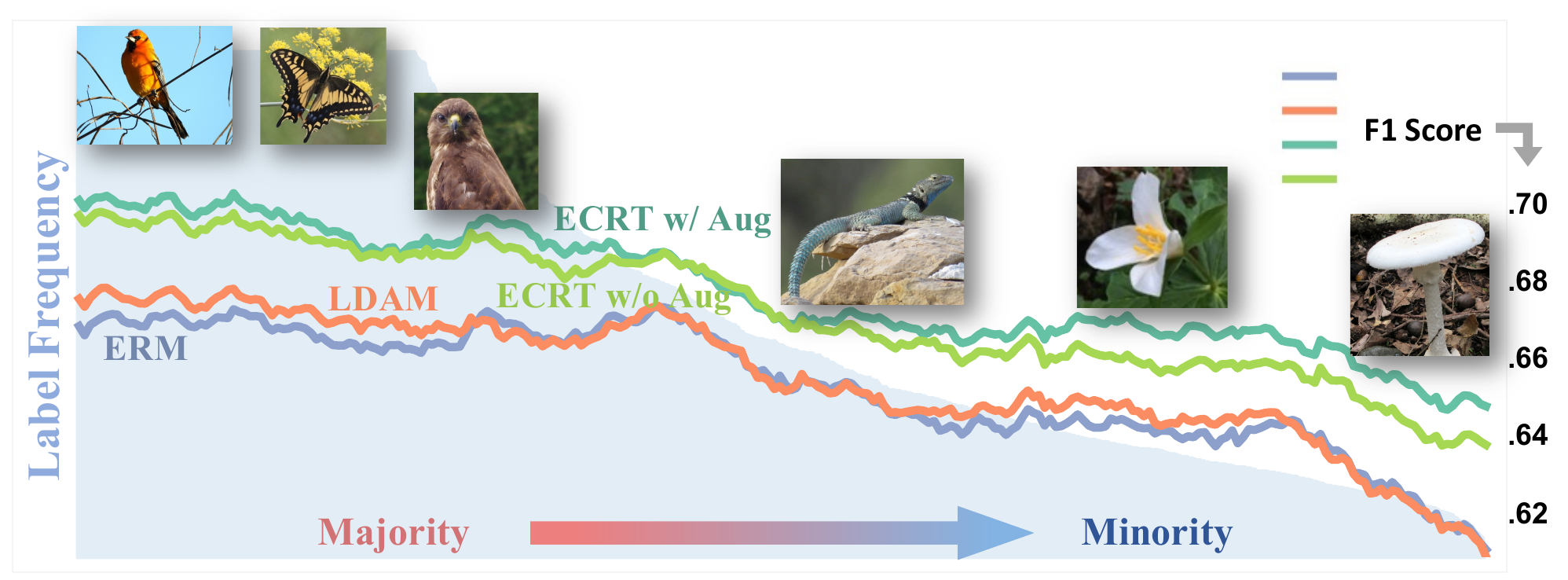}
\end{center}
\vspace{-.5em}
\caption{Energy-based causal representation transfer alleviates tail imbalance for natural image classification. The shaded region denotes label frequency in the \texttt{iNaturalist} data, with some representative images shown. Solid curves are for label-conditional F1 scores (higher is better) for the proposed {\it energy-based causal representation transfer} (ECRT), LDAM (state of the art) and the standard ERM. ECRT consistently outperforms the others, especially in the low-sample regime.
}
\label{fig:nat}
\vspace{-1.3em}
\end{figure}

Various research efforts have been directed to address the above issues, with class re-balancing being the most popular heuristic.
% re-establishing the group balance as the most popular heuristic.
The two most prominent directions in the category include statistical {\it resampling} and {\it reweighting}.
Resampling alters the exposure frequency during training, {\it i.e.}, more for the minorities and less so for the majorities \cite{drummond2003c4}.
Alternatively, reweighting directly amends the relative importance of each sample based on their class \cite{cui2019class}, sampling frequency \cite{botev2008efficient} and associated cost for misclassification \cite{elkan2001foundations}.
Recent developments also have considered class-sensitive and data-adaptive losses to more flexibly offset the imbalance \cite{cao2019learning, lin2017focal,menon2020long,kim2020m2m,tang2020long,kang2019decoupling}. 
While being intuitive and working reasonably well in practice, an important is that these approaches offer no protection against the over-fitting of minority instances (see Figure~\ref{fig:nat})\cite{xiu2020variational}, a fundamental obstacle towards better generalization.

To circumvent this key limitation, inductive biases are often solicited to impose strong constraints that suppress spurious correlations that hinder out-of-sample generalization.
A classical strategy is few-shot learning \cite{fei2006one, wang2020generalizing,hariharan2017low}, where the majority examples are used to train meta-predictors and transferable features, leaving only a few parameters to tune for the minority data.
In anomaly detection, methods such as one-class classification instead regard minority classes as outliers that do not always associate with stable, recognizable patterns \cite{moya1996network, ruff2018deep}.

Despite their relatively strong assumptions, these methods capitalize on their superior ability in generalizing in the low sample regime in empirical settings.

More recently, establishing causal invariance has emerged as a new, powerful learning principle for better generalization ability even under apparent distribution shifts \cite{rojas2018invariant}.
In contrast to standard empirical risk minimization (ERM) schemes, where the generalization to similar data distributions is considered, causally-inspired learning instead embraces robustness against potential perturbations \cite{arjovsky2019invariant}.
This is achieved via only attending to causally relevant features and associations postulated to be invariant under different settings \cite{peters2016causal}.
Specifically, contributions from spurious, unstable features are effectively blocked or attenuated.
Interestingly, compromise of performance can be expected in those models \cite{rothenhausler2018anchor}, as a direct consequence of discarding useful (but non-causal) correlations in exchange for better causal generalization. Recently, \cite{hyvarinen2019nonlinear} proposed non-parametric causal disentanglement of data representation to identify invariant relations. 

This work explores the advancement of imbalanced data learning via adopting causal perspectives, with the insight that the key algorithm can be reformulated as an energy-based contrastive learning procedure to drastically improve efficiency and flexibility.
In recognition of the limitations discussed above, we present {\it energy-based causal representation transfer} (ECRT): a novel imbalanced learning scheme that brings together ideas from causality, contrastive learning, energy modeling, data-augmentation and weakly-supervised learning, to address the identified weakness of existing solutions.
Our key contributions are:
($i$) a causal representation encoder informed by an invariant generative mechanism based on generalized contrastive learning;
($ii$) integrated data-augmentation and source representation regularization techniques exploiting feature independence to enrich minority representations that better balance the trade-off between utility and invariance; ($iii$) a key novelty is the derivation of an energy-based contrastive learning algorithm that greatly enhance model parallelism for large-label settings and extends generalized contrastive learning; 
and ($iv$) insightful discussions on the justifications for the use of the proposed approach. Our claims are supported by strong experimental evidence.

\vspace{-8pt}
\section{Preliminaries}
\begin{wrapfigure}{R}{0.5\textwidth}
\vspace{-2.2em}
\begin{center}
\includegraphics[width=.5\textwidth]{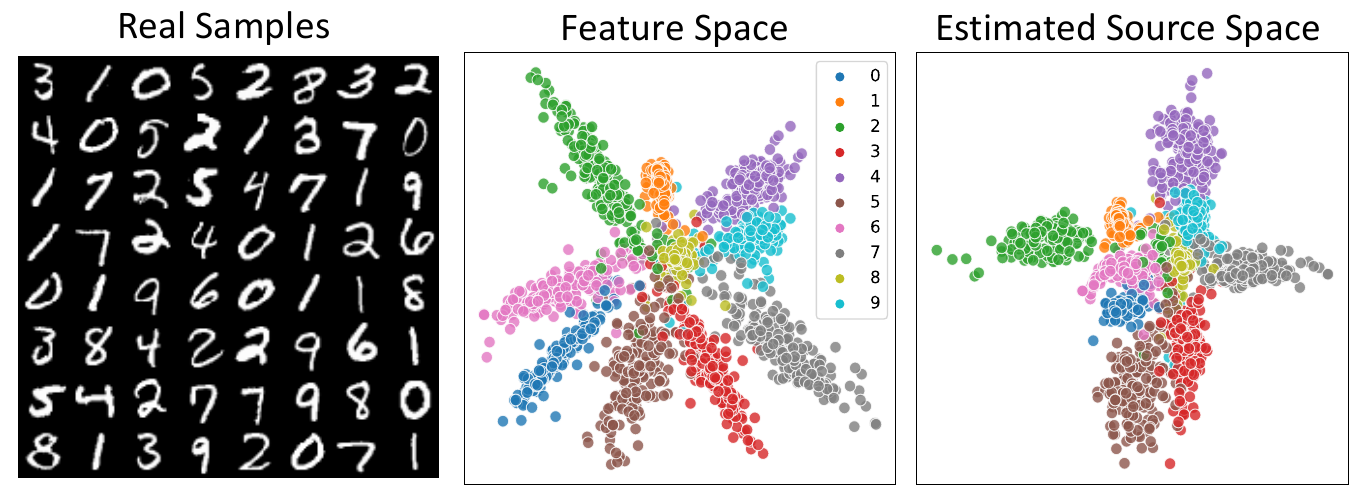}
\end{center}
\vspace{-1.2em}
\caption{Illustration of data space $\CX$, feature space $\CZ$ (predictive but entangled) and source space $\CS$ (independent, or disentangled) representations identified by ECRT for \texttt{MNIST}. 
}
\label{fig:mnist_sz_cmp}
\vspace{-1em}
% \end{figure}
\end{wrapfigure}

\vspace{-4pt}
{\bf Notation and problem definition.}
We use $\xv \in \BR^p$ to denote the input data and $\zv \in \BR^d$ for the predictive features extracted from $\xv$.
Let $y \in \{1,\ldots, M\}$ be the class label.
The number of training samples and those with label $m$ are denoted as $n$ and $n_m$, respectively.
We use $\EE[\cdot]$ to denote the expectation (average) of an empirical distribution, $\av_i$ to denote a vector associated with the $i$-th sample, and $[\av]_b$ to denote the $b$-th entry of vector $\av$.
For simplicity, we assume class $M$ is the minority class, {\it i.e.}, $n_m \gg n_M$, for $m \in \{1,\ldots, M-1\}$.
Throughout, we refer to $\CX, \CZ$ and $\CS$ as data, feature, and source spaces, respectively, as shown in Figure \ref{fig:mnist_sz_cmp}, and defined below.
Our goal is to accurately predict the label of minority instances with very limited training examples of it.
Generalization to multiple minority categories is straightforward.

{\bf Generalized contrastive learning and ICA} The proposed approach is based on a generalized form of {\it independent component analysis} (ICA), which addresses the inverse problem of signal {\it disentanglement} \cite{hyvarinen2000independent}.
Specifically, ICA decorrelates features $\Zv$ of the observed signal $\Xv$ into a source signal representation $\Sv = f_\psi(\Zv)$, where $f_\psi(\zv)$ is a smooth and invertible mapping known as the {\it de-mixing} function, and while assuming that the components of $\Sv$ are statistically independent, {\it i.e.}, with density $q(\sv) = \prod_j q_j([\sv]_j)$.
%, where $q(\sv)$ is commonly called the distribution of independent components. 
% More precisely, it tries to estimate a smooth, invertible de-mixing function $f(\zv)$, such that it disentangles $\Zv$ via $\Sv = f(\Zv)$. 
Notationally, we call $[\sv]_j$ the $j$-th {\it independent component} (IC) of $\Zv$.
% and $p_j([\sv]_j)$ its corresponding IC distribution.
% such $p_j([\sv]_j)$ {\it independent component distributions} (IC distributions). 
While {\it nonlinear ICA} (NICA)  is generally infeasible \cite{comon1994independent, hyvarinen1999nonlinear}, 
\cite{hyvarinen2019nonlinear} has recently proposed a setting in which the identification of NICA can be achieved, by requiring an additional auxiliary label $y$.
Specifically, NICA assumes that source signals are conditionally independent given $y$, {\it i.e.}, $q(\sv|y) = \prod_j q_j([\sv]_j|y)$, then $f_\psi(\zv)$ can be identified using {\it generalized contrastive learning} (GCL), whose implementation is detailed below.

% \iffalse
GCL solves a generalized regression problem in which a {\it critic function} predicts whether label $y$ and representation $\zv$ are correctly paired, {\it i.e.}, {\it congruent}. We call $(y_i, \zv_i)$ a congruent pair and $(y_j, \zv_i)$ an incongruent pair if $i\neq  j$.
Specifically, the critic is defined as $r_{\nu}(y,\zv) = \sum_{a=1}^d r_\nu^a(y,[\sv]_a)$, where $r_\nu^a(\cdot,\cdot)$ is a neural network with parameters $\nu$, whose inputs are the label $y$ and the $a$-th coordinate of $\sv=f_\psi(\zv)$, denoted as $[\sv]_a$.
Then, GCL optimizes the following objective:
\beq
\argmin_{f_\psi, r_\nu} \ \underbrace{\EE_{i}[ h(-r_\nu(y_i,\zv_i))] + \EE_{j \neq i} [ h(r_\nu(y_j,\zv_i)) ] }_{\CL_{\GCL}(f_\psi, r_\nu)}, 
\label{eq:gcl}
\eeq
where $h(r) = \log (1+\exp(r))$ is the softplus function.
\eqref{eq:gcl} seeks to optimize $f_\psi(\cdot)$ and $r_\nu(\cdot,\cdot)$ by maximizing the discriminative power to tell apart congruent and incongruent pairs.

In fact, an interesting result showed by \cite{hyvarinen2019nonlinear} revealed that maximizing the ability of the critic function $r_\nu(\cdot,\cdot)$ for correctly identifying matching pairs $(\zv, y)$ leads to the identification (up to univariate transformation) of $f_\psi(\cdot)$, such that the components of $\sv$ are conditionally independent given $y$.
See the Supplementary Material (SM) for a formal exposition.

\vspace{-8pt}
\section{Energy-based Causal Representation Transfer}
\vspace{-4pt}
In this section, we describe the construction of {\it energy-based causal representation transfer} (ECRT): a causally informed data transformation and augmentation procedure to improve learning with imbalanced datasets.
Our model assumes a shared causal data-generation procedure, which can be accurately identified by learning with the majority classes under assumed class-conditional representation independence.
We obtain decorrelated representations that facilitate data augmentation and efficient learning with minority classes.
% \new{We explicitly target} the inversion of this data generation mechanism, \new{to obtain a de-correlated representations for efficient learning and data augmentation.}

The proposed model consists of the following components:
($i$) a feature encoder module $\zv = e_{\theta}(\xv)$;
($ii$) two classification modules, $h_{\phi'}(\zv)$ and $h_{\phi}(\sv)$, for predicting label $y$ from features $\zv$ and sources $\sv$, respectively;
($iii$) a nonlinear ICA module for the de-mixing function $\sv = f_{\psi}(\zv)$;
($iv$) a critic function $r_\nu(y,\zv)$ for GCL;
and ($v$) a data augmentation module.
Further, $(\theta, \phi', \phi, \psi,\nu)$ denote the parameters of all the neural-network-based modules. 
Algorithm~\ref{alg:crt} outlines a general workflow, and below, we elaborate on our assumptions and detail the implementation of ECRT.

\begin{wrapfigure}[8]{R}{0.55\textwidth}
% \begin{figure}
\vspace{-3.8em}
\centering
\scalebox{0.9}{
\begin{minipage}{0.6\textwidth}
\begin{algorithm}[H]
   \caption{Energy-based Causal Representation Transfer.}
   \label{alg:crt}
\begin{algorithmic}
%\small
\STATE 1. Pre-train encoder and predictor:
\vspace{-.7em}
$$
e_{\theta}, h_{\phi'} \leftarrow \argmin \{ \EE[\ell(h(e(\xv)), y)] \}
$$
\vspace{-1.7em}
\STATE 2. NICA estimation with fixed $\zv=e_{\theta}(\xv)$:
\vspace{-.7em}
$$f_\psi, r_\nu \leftarrow \texttt{GCL}(\{(\zv, y)\}) \quad \text{\% Equation \eqref{eq:gcl}}$$ 
\vspace{-1.7em}
\STATE 3. Source space augmentation with fixed $\sv = f_\psi(\zv)$:
\vspace{-.7em}
$$(\tilde{\sv},y)\leftarrow 
\texttt{AUG}(\{\sv|y = M\})\quad \text{\% Equation \eqref{eq:np}}$$
\vspace{-1.7em}
\STATE 4. Minority predictor modeling with augmented source:
\vspace{-.7em}
$$h_{\phi} \leftarrow \argmin \{ \EE[\ell(h(\sv, y=M)] + \lambda \EE [\ell(h(\tilde{\sv},y=M)] \} 
\vspace{-.7em}
$$
\end{algorithmic}
\end{algorithm}
% }
\end{minipage}
}
\vspace{-1.5em}
\end{wrapfigure}
\begin{figure*}[t!]
\begin{center}
{
\begin{minipage}{.58\textwidth}
%\vspace{.7em}
% \vspace{-.5em}
\includegraphics[width=1.\textwidth]{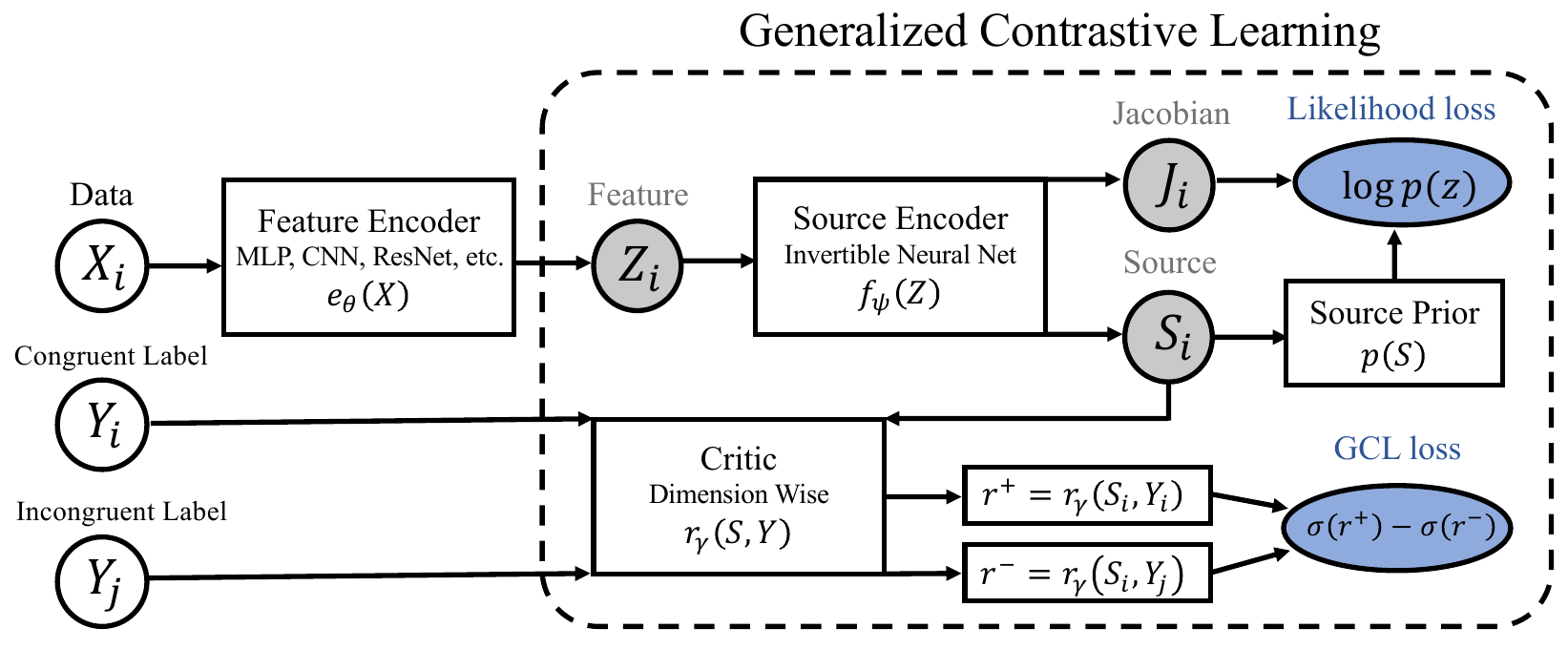}
\vspace{-1.5em}
\caption{
Source space estimation module of ECRT. We use GCL to identify the demixing function $\sv=f_{\psi}(\zv)$ via telling apart congruent \& incongruent pairs.
% \rh{Notation here needs to be consistent with the text. A ``fake'' label is never mentioned in the text.}
}
\label{fig:gcl}
\end{minipage}
\scalebox{1.}{
\begin{minipage}{0.4\textwidth}
\vspace{-.3em}
\includegraphics[width=1.\textwidth]{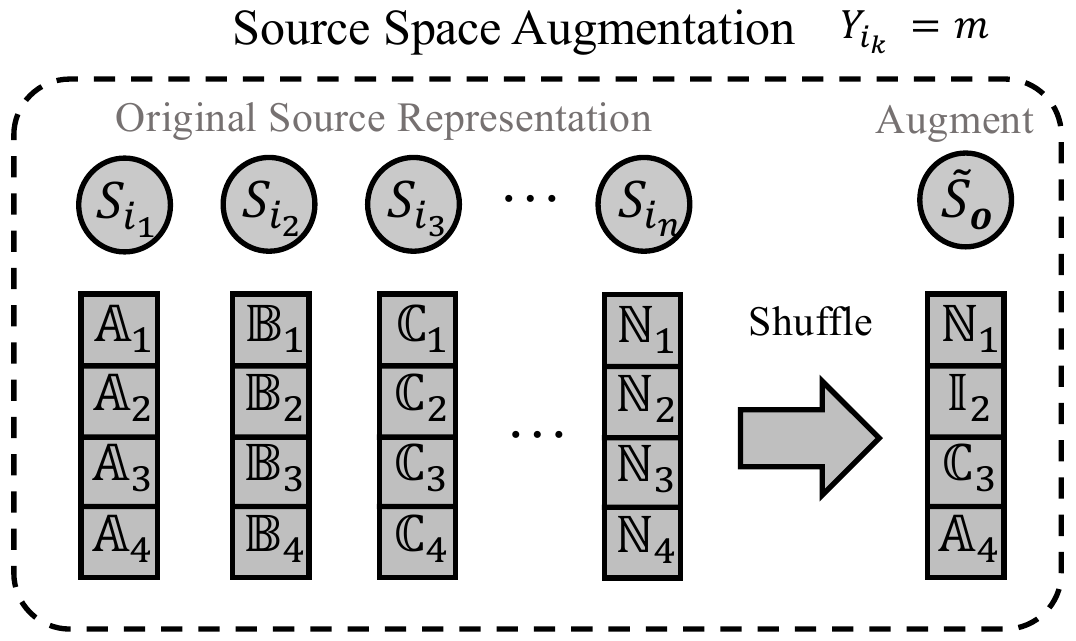}
\vspace{-1.5em}
\caption{
Non-parametric source space augmentation based on shuffling.
}
\label{fig:aug}
\end{minipage}
}
}
\end{center}
\vskip -.1in
\vspace{-1em}
\end{figure*}

\vspace{-10pt}
\subsection{Model assumptions}

To enable knowledge transfer across classes, we make the following assumptions: 

\begin{assumption}
Let $\zv$ be a sufficient statistic (features) of $\xv$ for predicting label $y$, all class conditional feature distributions $p(\zv|y)$ share a common ICA de-mixing function $f_\psi(\zv)$.
\label{thm:assump}
\end{assumption}

This implies there exists a smooth invertible function $f_\psi: \CZ \rightarrow \CS$, and a set of IC distributions $\{ q(\sv|y=m) \}_{m=1}^M$, that are linked to the conditional feature distributions $p(\zv|y=m)$ via $\Sv^{m} = f_\psi(\Zv^{m})$, where $\Sv^{m} \sim q(\sv|y=m)$ and $\Zv^{m} \sim p(\zv|y=m)$.
The subscript $m$ in $\Sv^{m}$ and $\Zv^{m}$ indicates $y=m$, in a slight abuse of notation. 

Importantly, $f_\psi^{-1}(\sv):\CS \rightarrow \CZ$ is the {\it invariant causal mechanism} underlying the features of observed data.
This specification, which is consistent with Assumption~\ref{thm:assump} enables the identification of the shared generating process, namely, the de-mixing function $f_\psi(\zv)$ that connects the likely dissimilar conditionals $\{p(\zv|y=m)\}_{m=1}^M$ via the source conditionals $\{q(\sv|y=m)\}_{m=1}^M$, as demonstrated in the context of NICA by \cite{hyvarinen2019nonlinear}

A shortcoming of Assumption \ref{thm:assump} is that the hypothesis supporting it is somewhat strong, untestable and may not hold in practice.
However, we argue that the structural constraints imposed on the model by the assumption via the source conditionals $q(\sv|y=m)$ and the invertibility of $f_\psi(\zv)$, restrict the search space of the otherwise over-flexible model space powered by neural networks. 
Further, the causal mechanism implied by $f_\psi^{-1}(\sv)$ enables an effective knowledge transfer mechanism across classes via $q(\sv|y=m)$.

\vspace{-8pt}
\subsection{Energy-based Causal Representation Transfer}
\label{sec:crt}

{\bf Encoder pre-training.}
To implement ECRT, we first find a good (reduced) feature representation of $\xv$ highly predictive of label $y$.
This can be achieved via supervised representation learning (see Figure \ref{fig:gcl}), which optimizes an encoder and predictor pair $(e_{\theta}(\xv), h_{\phi'}(\zv))$ to minimize the label prediction risk $\CL(\phi')\triangleq \EE [\ell(h_{\phi'}(e_\theta (\xv)),y)]$, where $\ell(\cdot,\cdot)$ is a suitable loss function, {\it e.g.}, cross-entropy, hinge loss, {\it etc.}
To avoid capturing {\it spurious} (non-generalizable) features that overfit the minority class, we advocate training only with majority samples at this stage; assuming that $M>2$.
Alternatively, one could also consider unsupervised feature extraction schemes, such as auto-encoders.
Further, we also recommend using statistical adjustments such as importance weighting, to reduce the impact of data imbalance.

{\bf De-mixing representation with GCL.}
After obtaining a good feature representation $\Zv = e_{\theta}(\Xv)$, we proceed to learn the de-mixing function $f_{\psi}(\zv)$, such that the coordinates of the source representation $\Sv = f_{\psi}(\Zv)$ are (approximately) independent given the label $y$.
This can be done by optimizing the GCL objective in \eqref{eq:gcl} with respect to the feature representation and label pairings $(y, \zv)$, adopting the {\it masked auto-regressive flow} (MAF) \cite{papamakarios2017masked} to model the smooth, invertible transformation $f_\psi(\zv)$, which allows efficient parallelization of the autoregressive architecture via {\it causal masking} \cite{chen2021variational}.
The procedure is outlined in Figure \ref{fig:gcl}.

{\bf Augmenting the minority.}
Inspired by \cite{teshima2020few}, we artificially augment the minority feature representations $\zv$ via random permutations in the source space ${\cal S}$, as shown in Figure \ref{fig:aug}.
Provided the assumed conditional independence of the sources, the features $\Zv^{m}$ corresponding to label $y=m$ are generated by $\Zv^{m} = f_\psi^{-1}(\Sv^{m})$, where each dimension in the source representation $[\Sv^{m}]_j \sim q([\sv]_j|y=m)$ are independently and {\it implicitly} sampled as described below.
% \rh{what is $q_j(s|y=m)$?}
% This implies that if we are able to sample from each marginal distribution given $y$, then we can synthesize new samples for $\Zv_y$.
%\rh{what is the difference between $\Zv$ and $\Zv_y$?}
Specifically, using the (estimated) de-mixing function $f_\psi(\zv)$, we can obtain an approximate empirical source distribution for each $y=m$, {\it i.e.}, $S^{m} \triangleq \{\sv_{i} = f_\psi(e_{\theta}(\xv_{i}))|y_i=m \} = \{ \sv_i \}_{i=1}^{n_m}$, where $S^{m}$ is a collection of $n_m$ samples of $q(\sv|y=m)$.
Then, we can draw new artificial samples $\tilde{\sv}^{m}\sim q(\sv|y=m)$ by randomly permuting the coordinates within elements $S^{m}$ independently via

\beq\label{eq:np}
\tilde{\sv}_{\ov}^{m} = ([\sv_{o_1}^{m}]_1, [\sv_{o_2}^{m}]_2, \cdots, [\sv_{o_d}^{m}]_d),
\eeq
where $\ov=(o_1, \cdots, o_d)$ is a random permutation of $(1,\ldots,n_m)$. 
Note we have used $\tilde{\sv}_{\ov}^m$ to emphasize that the source point is artificially created via permutation $\ov$.
We call this procedure {\it nonparametric augmentation} because it does not make distributional assumptions for $q(\sv|y=m)$.
However, below we will discuss its limitations and consider an alternative where a parametric form is assumed.
While it is tempting to refine the predictor $h_{\phi'}(\zv)$ with artificial features augmented via $\tilde{\zv}^{m} = f_\psi^{-1}(\tilde{\sv}^{m})$, in Section~\ref{sec:imp_crt} we will argue that it stands to benefit more from training a new predictor directly based on source representations, {\it i.e.}, $h_{\phi}(\sv)$, without the need for inverting $f_\psi(\zv)$.

{\bf Model refinement.}
Now we can leverage the augmented data to refine the prediction model.
For minority class $y=M$, we optimize the following objective
\beq\label{eq:laug}
\CL_{\text{AUG}}(\phi') = \CL(\phi') + \lambda ( \EE_{\tilde{\Zv}^M}[\ell(h_{\phi'}(\tilde{\zv}^{M}),M)] - \EE_{\Zv^M}[\ell(h_{\phi'}(\zv),M)]),
\eeq
where $\CL(\phi')$ is the loss used for pre-training.
Conceptually, \eqref{eq:laug} replaces a portion of the minority samples with augmentations.
The trade-off parameter $\lambda\in [0,1]$ encodes the relative confidence for trusting the artificially generated representations $\tilde{\Zv}^M$ obtained from $S^M$ for the minority label $y=M$.
Further, at this stage we found it's beneficial to fix the encoder module to prevent the de-mixing function to accommodate the changes in the encoder which in practice may cause instability during training.

{\bf Challenges with na\"ive implementation.}
We identify three major issues with na\"ively implemented ECRT, to be addressed in the section below: ($i$) {\it Representation conflict}: since the GCL solution is not unique, we do observe na\"ive GCL training drifts among viable source representations whose performance differ considerably, causing stability concerns; ($ii$) {\it Costly augmentation}: MAF inversions dominate the computation load during training, which becomes prohibitive in high dimensions; and ($iii$) {\it Gridding artifact}: a small minority sample size leads to pronounced augmentation bias when sampling nonparametrically via \eqref{eq:np}, manifested as a rectangular-shaped grid (see Figure \ref{fig:aug_cmp}).

\iffalse
\begin{itemize}
\vspace{-5pt}
\item {\bf Representation collapse.} While identifiable, GCL solution is not unique, and suboptimal ones degrade stability and performance.
%\rh{how does non-uniqueness affects identifiability?}
\vspace{-7pt}
\item {\bf Costly augmentation.} MAF inversions dominate the computation burden during training, \new{which becomes prohibitive in high dimensions};
%\rh{why is this a challenge? doe it become prohibitive?}
\vspace{-7pt}
\item {\bf Gridding artifact.} Small minority sample size leads to pronounced augmentation bias, manifested as rectangular-shaped grids (see Figure \ref{fig:aug_cmp}).
%\rh{need to define ``gridding''}
\end{itemize}
\fi

\vspace{-8pt}
\subsection{Improving causal representation transfer}
\label{sec:imp_crt}
\vspace{-5pt}

{\bf Energy-based GCL.} Our key insight to improve GCL comes from the fact that Equation (\ref{eq:gcl}) is essentially learning the density ratio between the joint and product of marginals, {\it i.e.}, $\frac{p(x,y)}{p(x)p(y)}$. This immediately reminds us the recent literature on contrastive {\it mutual information} (MI) estimators, such as InfoNCE \cite{poole2019variational}. In such works, a variational lower bound of MI is derived, and the algorithm optimizes a critic function using the positive samples from the joint distribution, and the negative samples from the product of the marginals. At their optimal value, these critics recover the density ratio or a transformation of it. Our development is based on the recent work of \cite{guo2021tight}, using an energy-perspective to improve contrastive learning. Specifically, we will be using a variant of the celebrated {\it Donsker-Varadhan} (DV) estimator \cite{donsker1983asymptotic}, and applied Fenchel duality trick to compute a solution \cite{fenchel1949conjugate,tao2019fenchel,dai2019exponential}. Specifically, the {\it Fenchel-Donsker-Varadhan} (FDV) estimator takes the following form:
\beq
I_{\text{FDV}} \triangleq \hat{I}_{\text{DV}}^K(\{\xv_i, \yv_i\})+ \frac{\sum_j \exp[(g_\theta(\xv_i, \yv_j)-g_\theta(\xv_i, \yv_i))/\tau ]}{\sum_j \exp[(\hat{g}_\theta(\xv_i, \yv_j)-\hat{g}_\theta(\xv_i, \yv_i))/\tau ]}+1,
\eeq
where $g_\theta(\xv_i, \yv_i)$ is our critic of interest and $\hat{I}^K_{\text{DV}}(\{\xv_k, \yv_k\}) = g(\xv_1, \yv_1)-\log(\sum_{k^\prime=1}^K \exp(g(\xv_1,\yv_k^\prime))/K)$ is the {\it Donsker-Varadhan} (DV) estimator \cite{donsker1983asymptotic} for the MI. 
% $y^+$ and $y^-$ represents randomly paired {\it positive} instances and {\it negative} instances.
Using the same parameterization used in GCL recovers the same causal identification property (see Appendix for details). Compared to the original GCL formulation, we are now using multiple negative samples instead of one, which greatly boosts learning efficiency \cite{gutmann2012bregman}. And this can be efficiently implemented with the {\it bilinear} critic trick \cite{chen2020simple,chen2021simpler,guo2021tight} so that all in-batch data can be used as negatives. In our context, it greatly boosts training efficiency when dealing with a large number of different classes. See Algorithm S1 in Appendix. 

% \noindent
{\bf Regularizing the data likelihood.}
Recall GCL solutions can only be identified up to an invertible transformation of each dimension \cite{hyvarinen2019nonlinear}, and the predictive performance of different valid GCL solutions can vary significantly. Empirically, we observe that a na\"ive implementation of GCL often leads to source representations that are densely packed (see Figure S1 in the SM).
This is undesirable, when decoding back to the feature space and making useful predictions, the neural network predictor will need to be expansive, {\it i.e.}, requiring a large Lipschitz constant, thereby sacrificing optimization stability and model generalization according to existing learning theory \cite{cortes1995support, vapnik2013nature}.

To encourage source representations that are less condensed, we consider a simple, intuitive strategy consisting of regularizing the source representation with the $\log$-likelihood in feature space.
This likelihood can be easily obtained with MAF using $\ell_{\FLOW}(f_\psi)$ defined in the SM. %\eqref{eq:flow}.
Following common practice, we set source prior $p(\sv)$ to the standard Gaussian, and optimize the following likelihood-regularized GCL objective:
\beq
\tilde{\CL}_{\GCL}(f_\psi, r_\nu) = \CL_{\GCL}(f_\psi, r_\nu) + \rho \CL_{\FLOW}(f_\psi), 
\label{eq:gcl_reg}
\eeq
where $\rho>0$ is the regularization strength.
Naturally, this regularization will encourage the global source representation identified by GCL to be more consistent with a Gaussian-shaped distribution. 

\begin{wrapfigure}[11]{r}{0.6\textwidth}
\vspace{-1.5em}
\begin{center}
\includegraphics[width=.58\textwidth]{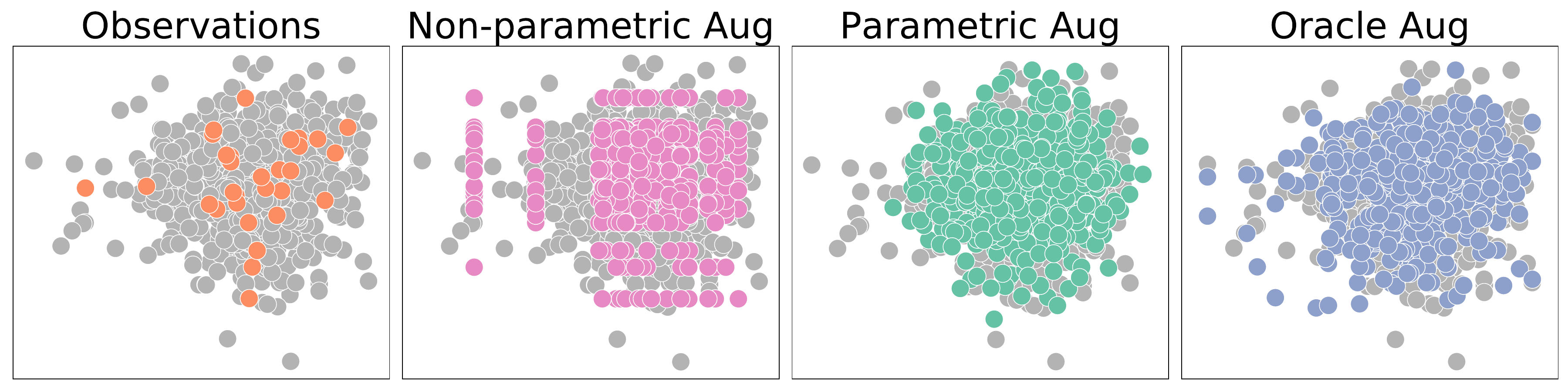}
\end{center}
\vskip -.2in
\caption{Comparison of different source augmentations overlaid on the ground-truth distribution (gray dots). A severe {\it gridding artifact} is observed in the low-sample regime for the nonparametric scheme, whereas the parametric augmentation closely matches the oracle in distribution.}
\label{fig:aug_cmp}
\end{wrapfigure}

An alternative interpretation for the likelihood-regularized objective in (\ref{eq:gcl_reg}) is that it can be understood as a relaxation to the conditional independence (Assumption \ref{thm:assump}).
To see this, recall the likelihood objective obtained from the {\it invertible neural networks} (INN)
alone attempts to map the source representations to be {\it unconditionally} independent, as opposed to the {\it conditional} independence assumed by NICA.
The regularized formulation (\ref{eq:gcl_reg}) provides a safe {\it ``fall-back''} mode in case Assumption \ref{thm:assump} is violated. This also motivates us to consider an important variant: making {multiple} class-dependent source priors, {\it i.e.}, $p^m(\sv)$ for each class label $m\in\{1,\cdots,M\}$ in (\ref{eq:gcl_reg}), whose parameters ({\it i.e.}, mean and variance) are jointly learned with other model parameters. 
Compared to the fully non-parametric objective (\ref{eq:gcl}), this strategy further encourages the source representations to be independent given the class labels, and it enables {\it parametric} data augmentation, {\it i.e.}, sampling from the parametric label priors instead of permuting the indices.
We refer to this variant as ECRT-MP, where MP stands for multiple priors. And similarly, ECRT-1P refers to the case when a single prior is used. We have found that ECRT-MP performs better in most cases, and consequently, ECRT means ECRT-MP by default.

% \begin{wrapfigure}[10]{r}{0.6\textwidth}
% % \begin{figure}
% \vspace{1.5em}
% \begin{center}
% % \includegraphics{figures/results/aug_cmp}
% \includegraphics[width=.58\textwidth]{figures/results/aug_cmp}
% \end{center}
% \vskip -.2in
% \caption{Comparison of different source augmentations overlaid on the ground-truth distribution (gray dots). A severe {\it gridding artifact} is observed in the low-sample regime for the nonparametric scheme, whereas the parametric augmentation closely matches the oracle in distribution.}
% \label{fig:aug_cmp}
% % \end{figure}
% \end{wrapfigure}

{\bf Modeling in the source space.}
Rather than modeling the predictor $h_{\phi'}(\zv)$ in the feature space $\CZ$, we advocate instead for building the predictor directly in the source space $\CS$, {\it i.e.}, modeling with $h_{\phi}(\sv)$.
This practice enjoys several benefits: ($i$) {\it Easy \& robust augmentation}: many designs of high-dimensional flows are asymmetric computationally, and inverting a MAF is not only $d$ times more costly than a forward pass, it is also numerically unstable at the boundary.
Direct modeling in the source space circumvents the difficulties associated with MAF inversions during data augmentation; 
% the costly and unstable MAF inversions to synthesize feature representations for data augmentation; 
($ii$) {\it Feature whitening}: 
% normalization has long been considered as an integral part of data analysis, supported by abundant empirical evidence.
the source representation identified by GCL is component-wise independent, and literature documents abundant empirical evidence that similar de-correlation based pre-processing, commonly known as {\it whitening}, benefits learning \cite{huang2018decorrelated, bansal2018can, jia2019orthogonal}.

{\bf Parametric augmentation.}
When the number of minority observations is scarce, the above non-parametric indices-shuffling augmentation suffers from the {\it gridding artifact} (Figure \ref{fig:aug_cmp}).
This artifact amplifies the augmentation bias in the low-sample regime.
To overcome this limitation, we empirically observe that the estimated class-conditional source distributions are usually Gaussian-like after the likelihood regularization (especially so when label conditional priors are used).
In these situations, a parametric augmentation that draws synthetic source samples from a Gaussian distribution matched to the empirical mean and variance of minority source representations is more efficient. 

\vspace{-8pt}
\subsection{Insights and remarks}

To better appreciate the gains and limitations expected from ECRT, we compile a few complementary arguments below, through the lens of very different perspectives.

\noindent {\bf Why causal augmentation works.}
It is helpful to understand the gains from ECRT's causal augmentation beyond the heuristic that permuting the ICs provides more training samples for the minority class. \cite{teshima2020few} considered a similar causal augmentation procedure for few-shot learning, and provided two major theoretical arguments: ($i$) the risk estimator based on the augmented examples is the uniformly minimum variance unbiased estimator given the accurate estimation of $f_{\psi}$ (see Theorem 1, \cite{teshima2020few}); and ($ii$) with high probability, the generalization gap can be bounded by the approximation error of $f_{\psi}$ (see Theorem 2, \cite{teshima2020few}). In the SM, we give arguments that our causal augments give the `best' label-conditional distribution estimate. 

% We are trying to fit the space here
% \new{why theory in few-shots can work here, how it relates to other part of this paper}

%A similar NICA-based data augmentation procedure has been considered in \cite{teshima2020few} for few-shot learning improvements, and the authors have provided t

% To complement the above augmentation-based theoretical results, here we provide alternative views to further justify GCL-based causal representation disentanglement.
% We only present informal, heuristic arguments in the main text, with more rigorous statements deferred to the SM.
% Our claims will be verified in the experiments.
% \rh{why not show the rigorous statements here? it seems preferable to to show rigor which can be seen as a contribution and leave the heuristics out, besides, the arguments below are too hand waving unless sharpened.}

{\bf Speedup from shared embedding.}
While for typical supervised learning tasks the generalization bound scale as $\CO(n^{-\frac{1}{2}})$, a superior rate of $\CO(n^{-\eta})$ where $\eta \in [\frac{1}{2}, 1]$ is possible, if there exists abundant data for an alternative, yet related task that shares the same feature embedding (see Theorem 3, \cite{robinson2020strength}).
Note $n$ refers to the size of labeled data directly related to the {\it strong} task of interest, in our case, prediction of minority labels.
% The general sufficient condition for that improved rate to be feasible is the existence of a shared mutual embedding for both the weak and strong tasks \cite{robinson2020strength}.
Our ECRT employs GCL to identify one such common embedding, {\it i.e.}, the source space, using the majority examples, and consequently, improves predictions on the minority class. 
% the main task of predicting minority labels,.

{\bf Representation whitening.}
Our ECRT causally disentangles representation \cite{siddharth2017learning, tokmakov2019learning} via de-correlating the representations conditionally. 
Extensive empirical evidence has pointed to the fact that such representation de-correlation, more commonly known as data whitening \cite{kessy2018optimal}, is expected to considerably improve learning efficiency \cite{cogswell2015reducing}.
This benefit has been attributed to the better conditioning of the Fisher information matrix for gradient-based optimization \cite{desjardins2015natural}, thus rectifying second-order curvatures to accelerate convergence. 
Our source space modeling explicitly separates the task of representation disentanglement, and in turn, helps the prediction network to focus on its primary goal.

{\bf Potential limitations.}
% We caution there is more nuisance involved in the discussion.
The setting considered by ECRT is restrictive in that it precludes the learning of useful, yet non-transferable features predictive of the minority labels.
For instance, there might be a feature unique to the minority class.
% \rh{what is a signature?}
However, since the de-mixer $f_{\psi}(\zv)$ is only trained on the majority domains absent of this feature, it can not be accounted for by the ECRT model.
This is a key limitation of causally inspired models, in that they are often too conservative for only retaining the invariant features, promoting cross-domain generalization at the cost of within-domain performance degradation \cite{rothenhausler2018anchor}.

% Formulating a hybrid solution that combines the wisdom of other imbalanced learning strategies promises to overcome this limitation, and we leave it for future investigations. 

\vspace{-8pt}
\section{Related Work}
\vspace{-4pt}
\noindent {\bf Causal invariance and representation learning.} 
A major school of considerations for building robust machine learning models is to stipulate invariant causal relations across environments \cite{scholkopf2012causal,lu2020reconsidering,chen2021wiener}, such that one hopes to safely extrapolate beyond the training scenario \cite{buhlmann2018invariance}. We broadly categorize such efforts into two streams, namely predictive causal models and generative causal models. Prominent examples from the first category include ICP \cite{peters2016causal, heinze2018invariant} and IRM \cite{arjovsky2019invariant}, which highlight the identification of invariant representation and causal relations via penalizing environmental heterogeneity. Our solution pertains to the second category, where the data distribution shifts across environments can be tethered by an invariant generation procedure \cite{hyvarinen2019nonlinear, khemakhem2020variational}. This work exploits the recovered causally invariant source representation to improve learning efficiency and mitigate the sample scarcity of minority labels in imbalanced sets.

\begin{figure*}[t!]
\begin{minipage}{.3\textwidth}
\vspace{-.6em}
\begin{center}
\includegraphics[width=1.\textwidth]{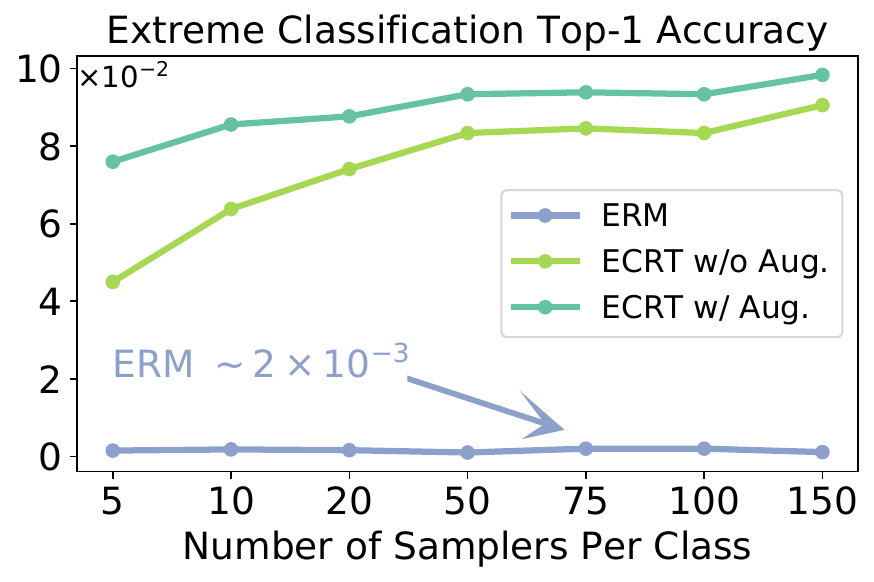}
\end{center}
\vspace{-1.4em}
\caption{Extreme classification of $1k$ labels. ERM performs slightly better than random guess, while ECRT works significantly better. \label{fig:extreme_top1}}
\end{minipage}
\hspace{2pt}
\begin{minipage}{.3\textwidth}
\vspace{-.6em}
\begin{center}
\includegraphics[width=1.\textwidth]{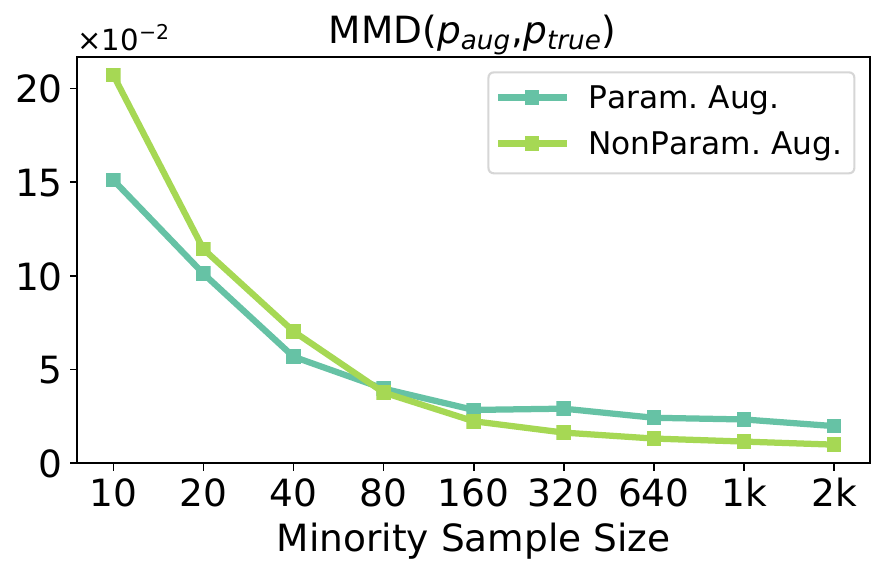}
\end{center}
\vspace{-1.6em}
\caption{Comparison of different causal augmentations, lower is better. Parametric augmentation is more efficient with small samples. \label{fig:aug_mmd}}
\end{minipage}
\hspace{2pt}
\begin{minipage}{.33\textwidth}
\vspace{-0.8 em}
\begin{center}
\includegraphics[width=1.\textwidth]{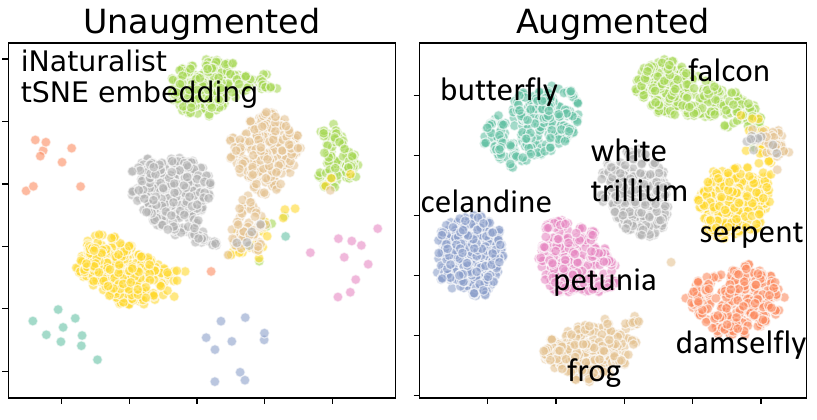}
\end{center}
\vspace{-0.5 em}
\caption{ECRT source representation trained on \texttt{iNaturalist}, visualized using tSNE embedding in two dimensions for eight random categories. \label{fig:inat_tsne}}
\end{minipage}
\vspace{-1em}
\end{figure*}

% \vspace{-3pt}
{\bf Learning with imbalanced data.} 
Resolving data imbalance is a heavily investigated topic \cite{liu2019large}. Standard sampling and weight adjustments suffer from caveats such as introducing bias and information loss. Due to these concerns, recent literature has actively explored adaptive strategies, such as redundancy-adjusted balancing weights \cite{cui2019class}, and the theoretically grounded class-size adapted margins\cite{cao2019learning}. Similar to {\it boosting} \cite{freund1997decision}, adaptive weights are designed to prioritize the learning of less well-classified examples \cite{lin2017focal} while excluding apparent outliers \cite{li2019gradient}. Much related to our setting are the {\it meta-learning} scenarios \cite{vinyals2016matching, finn2017model, wang2017learning}, where the model tries to generalize \& repurpose the knowledge learned from majority classes to minority predictions; and also the augmentation-based schemes to restore class balance \cite{mariani2018bagan, mullick2019generative}. 

% \vspace{-3pt}
{\bf Data augmentation.} Due to its exceptional effectiveness, augmentation schemes are widely adopted in practical applications \cite{shorten2019survey}. These augmentation strategies are built on known invariant transformations, {\it e.g.}, rotation, scaling, noise corruption, {\it etc}. \cite{szegedy2016rethinking}, simple interpolation heuristics \cite{chawla2002smote, he2008adasyn}, and more recently towards fully auto-mated procedures \cite{cubuk2019autoaugment}. Notably, recent trend in data augmentation highlights robust learning against adversarially crafted inputs \cite{goodfellow2014explaining} and generative augmentation procedures that compose realistic artificial samples \cite{antoniou2017data}.  

{\bf Domain adaption and causal mechanism transfer.} Closest to our contribution is {\it causal mechanism transfer} (CMT) \cite{teshima2020few}, which focused on addressing few-shot learning for continuous regression. We note a few key differences to our work: ($i$) CMT focused on domain adaptation and does not address classification; and ($ii$) it bundles $(\xv, y)$ for NICA which necessitates the flow inversion for sample augmentation. Grounded in the setting of imbalanced data learning, our ECRT extends applications and advocates source space modeling to simplify and improve causal augmentation. It also features likelihood regularization to enhance representation regularity. We also offer new insights to justify the use of NICA-based augmentation, complementing the analysis from CMT by \cite{teshima2020few}. 

{\bf Energy-based modeling for representation learning.} There is growing recognition that the energy perspective is integral to representation learning \cite{lecun2006tutorial,poole2019variational, tao2019variational, guo2021tight}. In the lens of energy based modeling, the distribution of data is characterized as an (unnormalized) energy function \cite{arbel2020generalized}. Optimization of the energy function is often considered challenging \cite{tao2019fenchel} and contrastive techniques have been proven effective \cite{hinton2002training,gutmann2010noise}. Our work is a generalization of \cite{hyvarinen2019nonlinear} that disentangles representations from an energy perspective.
Interesting comparison can be made to \cite{khosla2020supervised}, whose training objective bears resemblance to our ECRT. But the specific designs used by ECRT in the network architecture and scoring function allows ECRT to provably disentangle the feature representations and consequently capture causality for exploitation.

\vspace{-8pt}
\section{Experiments}
\vspace{-4pt}
\noindent To validate the utility of our model, we consider a wide range of (semi)-synthetic and real-world tasks experimentally. All experiments are implemented with PyTorch, and our code is available from \url{https://github.com/ZidiXiu/ECRT}. More details of our setup \& additional analyses are deferred to the SM Sections C-E.

\vspace{-8pt}
\subsection{Experimental setup}
\label{sec:expr}

\noindent{\bf Baselines.} The following competing baselines are considered to benchmark the proposed solution: $(i)$ Empirical risk minimization (ERM), a na\"ive baseline with no adjustment; ($ii$) Importance-weighting (IW) \cite{byrd2019effect}, a class-weight balanced training loss; ($iii$) Generative adversarial augmentation (GAN) \cite{antoniou2017data}, synthetic augmentations from adversarially-trained sampler; ($iv$) Virtual adversarial training (VAT) \cite{miyato2018virtual}, robustness regularization with virtual perturbations; ($v$) FOCAL loss \cite{lin2017focal}, cost-sensitive adaptive weighting; ($vi$) Label-distribution-aware margin loss (LDAM) \cite{cao2019learning}, margin-optimal class weights. All baselines are tuned for best performance for {\sc top-1} accuracy on validation.

{\bf Evaluation metrics \& setup.} We consider the following metrics to quantitatively assess performance: ($i$) {\it negative $\log$-likelihood} (NLL); ($ii$) F1 score; ($iii$) {\sc Top-$k$} accuracy ($k=1,5$). Following the classical evaluation setup for imbalanced data learning, we learn on an imbalanced training set and report performance on a balanced validation set.

\noindent {\bf Toy model.} We sample 2D standard Gaussians with different means and variances as source representation $\sv$ for each label class, which is then distorted by random affine and H\'enon transformations $z_1 = 1 - 1.4\, \tilde{s}_1^2 + \tilde{s}_2, z_2 = 0.3 \, \tilde{s}_1$, to induce real-world-like complex association structures. See the SM Sec. D for a detailed description and visualizations. 

\begin{figure*}[t!]
\begin{minipage}{.32\textwidth}
\vspace{-.3em}
\centering
\includegraphics[width=1.\textwidth]{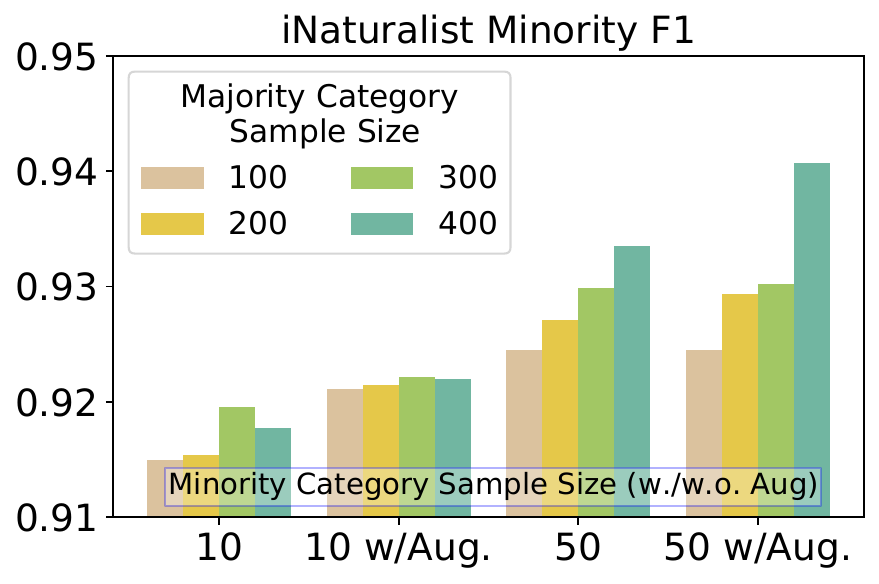}
\vspace{-1.5em}
\caption{Performance comparison with and without augmentation. Augmented solutions improve the effective sample size. \label{fig:inat_aug_eff}}
\end{minipage}
\hspace{1pt}
\begin{minipage}{.32\textwidth}
\vspace{-1.1em}
\centering
\includegraphics[width=1.\textwidth]{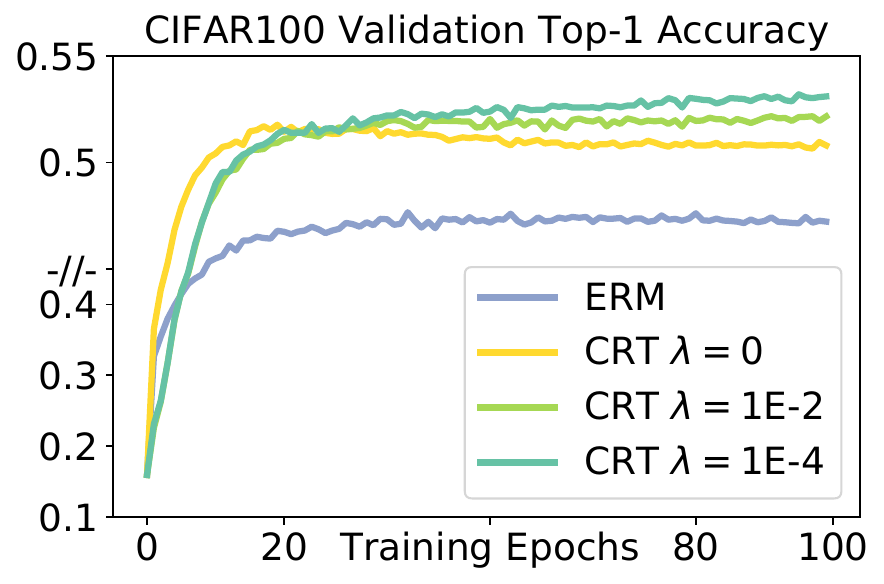}
\vspace{-1.55em}
\caption{Comparison of learning dynamics. ECRT enables both faster learning and better model predictions. \label{fig:cifar_learn}}
\end{minipage}
\hspace{1pt}
\begin{minipage}{.33\textwidth}
\vspace{-0.9em}
\centering
\includegraphics[width=1.\textwidth]{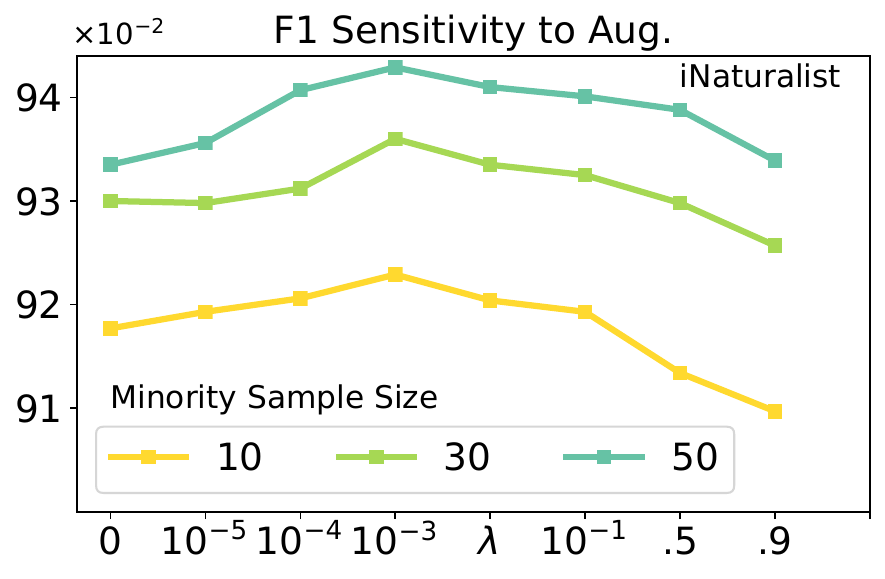}
\vspace{-1.7em}
\caption{Sensitivity analysis of augmentation strength $\lambda$. Smaller minority sample sizes are more sensitive to the augmentation. \label{fig:inat_aug_curve}}
\end{minipage}
\vspace{-.5em}
\end{figure*}

\begin{table}[t!]
\caption{Comparison of performance on real-world datasets ($\uparrow$ higher is better, $\downarrow$ lower is better). \label{tab:semi} }
\vspace{-.7em}
\setlength{\tabcolsep}{3pt}
\renewcommand{\arraystretch}{1.}
\begin{center}
\begin{sc}
\resizebox{\columnwidth}{!}{
\begin{tabular}{cccc@{\hskip 1em}ccc@{\hskip 1em}ccc@{\hskip 1em}cc}
\toprule
& \multicolumn{3}{c}{\texttt{CIFAR100}} & \multicolumn{3}{c}{\texttt{iNaturalist}} & \multicolumn{3}{c}{\texttt{TinyImagenet}}  & \multicolumn{2}{c}{\texttt{ArXiv}} \\
% \midrule
 & Top-1$\uparrow$ & Top-5$\uparrow$ & NLL$\downarrow$ & Top-1$\uparrow$ & Top-5$\uparrow$ & NLL$\downarrow$ & Top-1$\uparrow$ & Top-5$\uparrow$ & NLL$\downarrow$ & \multicolumn{1}{c}{Acc$\uparrow$} &\multicolumn{1}{c}{NLL$\downarrow$} \\ 
 \midrule
ERM & 49.29 & 78.22 & 2.95 & 66.73 & 87.86 & 1.70 & 58.52 & 79.01 & 3.22 & 44.64 & {\bf 0.0407} \\
IW  & 43.97 & 68.96 & 3.89 & 67.63 & 88.94 & 1.66 & 60.50 & 80.23 & 2.92 & 46.02 & 0.0477 \\
GAN & 47.64 & 78.53 & 2.69 & 67.40 & 87.00 & 1.82 & 60.69 & 80.91 & 2.33 & 45.42 & {\bf 0.0407} \\
VAT & 46.47 & 74.23 & 3.19 & 67.06 & 87.39 & 1.90 & 59.69 & 82.05 & 2.42 & 45.82 & 0.0474 \\
FOCAL&43.32 & 74.39 & 2.89 & 66.63 & 88.45 & 1.59 & 58.27 & 79.39 & 2.59 & 46.01 & 0.0416\\
LDAM &50.46 & 74.39 & 2.18 & 67.39 & 87.13 & 4.00 & 58.18 & 82.51 & 2.15 & 45.04 & 0.0450\\
[5pt]
ECRT-1P & 52.31 & 81.14 & {\bf 1.98} & 68.38 & 88.13 & 1.48 & 62.46 & 83.50 & {\bf 1.79} & \multirow{2}{*}{\bf 48.33} & \multirow{2}{*}{0.0434}\\
ECRT-MP & {\bf 53.00} & {\bf 81.99} & { 2.31} & {\bf 69.01} & {\bf 90.01} & {\bf 1.23} & {\bf 64.40} & {\bf 84.54} & { 1.94} & \\ 
% CRT FDV &  &  &  &  &  &  &  &  &  & \\
\bottomrule
\end{tabular}
}
\end{sc}
% \end{small}
\end{center}
\end{table}

\noindent{\bf Real-world datasets.} We consider the following semi-synthetic and real datasets: ($i$) Imbalanced \texttt{MNIST} and \texttt{CIFAR100}: standard image classification tasks with artificially created step-imbalanced following \cite{cao2019learning}; ($ii$) Imbalanced \texttt{TinyImageNet} \cite{tiny2017}: a scaled-down version of the classic natural image dataset \texttt{ImageNet}, comprised of $200$ classes, $500$ samples per class and $10k$ validation images, with different simulated imbalances applied; ($iii$) \texttt{ iNaturalist 2019} \cite{van2018inaturalist}: a challenging task for image classification in the wild comprised of $26k$ training and $3k$ validation images labeled with $1k$ classes, with a skewed distribution in label frequency; ($iv$) \texttt{arXiv} abstracts, imbalanced multi-label prediction of paper categories with $160k$ samples and $152$ classes. 

% \vspace{-3pt}
{\bf Preprocessing, model architecture, and tuning.} We use pre-trained models to extract vectorized representations for big complex datasets: ResNet \cite{he2016deep} for image models of \texttt{TinyImageNet}\footnote{\url{https://download.pytorch.org/models/resnet18-5c106cde.pth}}, Inception V3 for \texttt{iNaturalist}\footnote{\url{https://download.pytorch.org/models/inception_v3_google-1a9a5a14.pth}},  and BERT \cite{devlin2019bert}\footnote{\url{https://github.com/allenai/scibert}} for the \texttt{arXiv} \cite{arxiv2020} language model. These pre-trained representations are used as raw feature inputs, subsequently fed into a fully-connected multi-layer perceptrons (MLP) for source space encoding. For small image models ({\it i.e.}, \texttt{MNIST}, \texttt{CIFAR100}) we directly train a CNN from scratch for feature extraction.
We used a random $8/2$ split for training and validation, and applied Adam optimizers for training. We rely on the best out-of-sample cross-entropy and GCL loss for hyperparameter tuning.

{\bf ECRT efficiency.} We first examine the dynamics of efficiency gains from ECRT using a toy model, with the results summarized in Figure \ref{fig:inat_aug_eff}. Consistent with the shared embedding perspective, increasing the majority sample size improves the minority accuracy. Causal augmentation consistently improves performance, and it is most effective in the low-sample regime, offering more than $4 \times$ boosts in effective minority sample size. In Figure \ref{fig:cifar_learn}, we compare modeling with feature and source representations, respectively. Source space modeling speeds up learning, in addition to the modest accuracy gains. Augmented training is slightly slower, but eventually converges to a better solution. Interestingly, the major performance gain originates from the GCL-based source space modeling we proposed, and to a lesser extent, from the augmentation perspective presented by \cite{teshima2020few}.

{\bf Ablations on augmentation.} We examined the contributions from augmentation as we vary the augmentation strength $\lambda$ (Figure \ref{fig:inat_aug_curve}). A relatively small $\lambda$ already yields improvement, but when the minority size is small, stronger augmentation degrades performance. This implies the augmentation gains are more likely to be originated from the exposure to the diversity of synthetic augmenting, and the accuracy of augmented distribution is limited by the imperfectness of empirical estimation ({\it e.g.}, limited sample size). In Figure \ref{fig:aug_mmd} we compare the MMD distance \cite{gretton2012kernel} between the parametric and nonparametric augmentation outputs and the ground-truth, confirming the improved efficiency from the ECRT parametric augmentation in the low-sample regime.  

{\bf Ablation on different model variants.}
We further compare the model performance with feature encoder trained with both majority and minority. (Table \ref{tab:semi} used feature encoder trained only majority classes.) We have \texttt{Cifar100} {\sc top-1}=51.79, {\sc top-5}=81.02, {\sc NLL}=2.03. This is slightly worse than the majority-only result but still outperformed other competing solutions. 
We caution, whether including minority examples in the pre-training of feature encoder differs case-by-case. They may not affect performance at all (majority dominant), improve performance (predictive features consistent with those used by majority) and completely devastating (containing spurious features, overfit).

{\bf Extreme classification.} A related challenge of interest is extreme classification \cite{choromanska2013extreme}, where the label size is huge but there is only a handful of samples pertaining to each label class, This means there is no clear majority class. In Figure \ref{fig:extreme_top1}, we show the proposed ECRT also fares very well in this scenario, while standard ERM struggles (slightly better than random guess). This result evidences that ECRT also efficiently extracts generalizable information from an abundance of different class labels.

{\bf Imbalanced data learning.} In Table \ref{tab:semi} we compare the performance of ECRT to other competing solutions, with each baseline carefully tuned to ensure fairness. Minor discrepancies compared to results reported by prior literature are explained in the SM Sec. E, and in general our implementation performs slightly better. Overall, ECRT-based solutions consistently lead the performance chart, with the label-prior variant solidly outperforming vanilla ECRT and other competitors in most categories. While showing varying degrees of success, FOCAL and LDAM failed to establish dominance compared to ERM. GAN and VAT also verified the effectiveness of augmentation and adversarial perturbations in the imbalanced setting.
We also compare the overall F1-score between different methods on \texttt{CIFAR100} dataset: ERM: 0.439, SMOTE \cite{chawla2002smote}: 0.444, FOCAL:0.391, LDAM: 0.408, and ECRT: 0.482. FOCAL and LDAM gave very poor F1 scores in this case, even worse than the ERM baseline. SMOTE improved ERM, but the largest gain is obtained by ECRT. Figure \ref{fig:nat} compares performance of different learning schemes condition on different label sizes using the F1 score on the \texttt{iNaturalist} data. 
In Figure \ref{fig:inat_tsne} we show the ECRT learned representations, with and without augmentation, using tSNE embeddings.

\vspace{-8pt}
\section{Conclusion}
\vspace{-4pt}
\noindent This paper developed a novel learning scheme for imbalanced data learning. Leveraging a causal perspective, our solution cuts through the apparent data heterogeneity and identifies a shared, invariant, disentangled source representation, using only majority samples. We demonstrate this source representation modeling enables more efficient learning and allows principled augmentation. Importantly, we bridge the research between contrastive representation learning and energy modeling to causal machine learning, which constructs promising directions for future research.

\vspace{-8pt}
\section*{Acknowledgements}
\vspace{-4pt}
This research was supported in part by NIH/NIDDK R01-DK123062, NIH/NIBIB R01-EB025020, NIH/NINDS 1R61NS120246, DARPA, DOE, ONR and NSF. J. Chen was partially supported by Shanghai Municipal Science and Technology Major Project (No.2018SHZDZX01) and National Key R\&D Program of China (No.2018AAA0100303).
This work used the Extreme Science and Engineering Discovery Environment (XSEDE), which is supported by National Science Foundation grant number ACI-1548562 \cite{towns2014xsede}.
This work used the Extreme Science and Engineering Discovery Environment (XSEDE) PSC Bridges-2 and SDSC Expanse at the service-provider through allocation TG-ELE200002 and TG-CIS210044.

% \clearpage
{\small
\bibliography{crt}
\bibliographystyle{plain}
}

\end{document}

\onecolumn 
\title{Supercharging Imbalanced Data Learning With Energy-based \\Contrastive Representation Transfer (Supplementary Material)}

\author{Zidi Xiu, Junya Chen, Benjamin Goldstein, Ricardo Henao, Lawrence Carin, Chenyang Tao\\
Duke University\\
}
\maketitle

% \beginsupplement

% <==========================================================
%\appendix
% \section*{Supplementary Material to `Supercharging Imblanced Data Learning With Causal Representation Transfer'}
% \addcontentsline{toc}{section}{Appendix} 
% \addcontentsline{toc}{section}{Appendix}
% \part{Appendix}
\hypersetup{colorlinks=true, linkcolor=black}
\tableofcontents
\appendix
\hypersetup{linkcolor=red,urlcolor=blue}

\section{Theoretical Support}

Here we summarize some theories from literature that supports the development of this paper by making it self-contained. Attempts have been made to unify the notations, making them consistent with our paper, and also drop some contents from the original presentations that not directly relevant in this context. 

\subsection{Nonlinear ICA with auxiliary variables}

The following theory lists the technical conditions required for the identification of conditional nonlinear ICA model, based on which  our work was built. 

\begin{defn}[Conditionally exponential of order $k$] 
\label{defn:cond_exp}
A random variable (independent component) $[\sv]_i$ is conditionally exponential of order $k$ given random vector $\cv$ if its conditional pdf can be given in the form 
\beq
p([\sv]_i|\cv) = \frac{Q_i([\sv]_i)}{Z_i(\cv)} \exp\left[ \sum_{j=1}^k \tilde{q}_{ij}([\sv]_i) \lambda_{ij}(\cv) \right]
\eeq
almost everywhere in the support of $\cv$, with $\tilde{q}_{ij}, \lambda_{ij}, Q_i$ and $Z_i$ scalar-valued functions. The sufficient statistics $\tilde{q}_{ij}$ are assumed linearly independent. 
\end{defn}

\newcommand{\Lv}{\bs{L}}
\begin{thm}[Theorem 3, \cite{hyvarinen2019nonlinear}, identification of Nonlinear ICA] Assume ($i$) the data follows the nonlinear ICA model with the conditional independence $q(\sv|\cv) = \prod_j q_j([\sv]_j|\cv)$; ($ii$) Each $[\sv]_j$ is conditionally exponential given $\cv$ (Definition \ref{defn:cond_exp}); ($iii$) There exist $n k + 1$ points $\cv_0, \cdots, \cv_{nk}$, such that the following matrix of size $nk \times nk$
\beq
\tilde{\Lv} = \left( 
\begin{array}{ccc}
\lambda_{11}(\cv_1) - \lambda_{11}(\cv_0) & \cdots & \lambda_{11}(\cv_{nk}) - \lambda_{11}(\cv_0) \\
\lambda_{nk}(\cv_1) - \lambda_{nk}(\cv_0) & \cdots & \lambda_{nk}(\cv_{nk}) - \lambda_{nk}(\cv_0)
\end{array}
\right)
\eeq
is invertible; ($iv$) nonlinear Logistic regression system Eqn (1) is trained using functions with universal approximation capacity. Then in the limit of infinite data, $f(\zv)$ provides a consistent estimator of the nonlinear ICA model, up to a linear transformation of point-wise scalar functions of the independent components. 
\end{thm}

\subsection{Variance and generalization bound}

The following theories explore the consequence of training using only nonparametric causal augmentation. First we define the risk estimators.

\begin{defn}
Let $\tilde{S}$ be the non-parametric source augmentation defined in Eqn (3) main text, $\ell(\cdot)$ be the loss function, $g(\zv)$ be the hypothesis function. We define the risk $R$ and causally augmented risk  estimator $\breve{R}$ wrt $g$ respectively as 
\beq
R(g) \triangleq \EE_Z[\ell(g(Z))], \breve{R}(g) \triangleq \EE_{\tilde{S}}[\ell(g(\hat{f}^{-1}(\tilde{S})))], 
\eeq
where $\hat{f}$ is the estimated causal de-mixing function. 
\end{defn}

The following theorem revealed that assuming perfect knowledge of de-mixing function $f$, the causally augmented risk estimator is optimal.

\begin{prop}
Assuming $\hat{f} = f$, and let $\ell(h(\xv), y)$ be the classification loss for predictor $h\in\CH$. Let $\hat{R}(h) = \sum_m w_m \hat{R}_m(h)$ be an estimator for $R(h) = \EE[\ell(h(\xv), y)]$, such that $\hat{R}(h)$ is an unbiased estimator for $R_m(h) = \EE_{y=m}[\ell(h(\xv),y)]$. Then $\tilde{R}(h) = \sum_m w_m \tilde{R}_m(h)$, where $\tilde{R}_m(h) \triangleq \EE_{y_i=m}[\ell(h(\tilde{\xv}_i),y_i)]$ is the minimum variance unbiased estimator among all $\hat{R}(h)$.
\end{prop}

\begin{proof}
This is a direct consequence of Theorem \ref{thm:minvar}. If $\tilde{R}(h)$ is not the minimal variance estimator, then at least one of $\tilde{R}_m(h)$ is not optimal, which contradicts Theorem \ref{thm:minvar}. 
\end{proof}

\newcommand{\CG}{\mathcal{G}}

\begin{thm}[Theorem 1, \cite{teshima2020few}, minimum variance property] Assuming $\hat{f} = f$. Then for each $g \in \CG$, the causal augmented risk estimator $\tilde{R}(g)$ is the uniformly minimum variance unbiased estimator of $R(g)$, i.e., $\EE[\breve{R}(g)]=R(g)$ and for any unbiased estimator $\breve{R}$ of $R(g)$ ({\it i.e.}, $\EE[\breve{R}(g)]=R(g)$),
\beq
\text{Var}[\tilde{R}(g)] \leq \text{Var}[\breve{R}(g)]. 
\eeq
\label{thm:minvar}
\end{thm}

Since we are bound to have estimation errors, the next theorem establishes the generalization bounds wrt such errors. 

\begin{thm}[Theorem 2, \cite{teshima2020few}, excess risk bound]
Let $\breve{g} = \argmin \breve{R}$ and $g^* = \argmin R(g)$, then under appropriate assumptions (Assumptionss 1-8 in \cite{teshima2020few}), for arbitrary $\delta, \delta' \in (0,1)$, we have probability at least $1-(\delta + \delta')$, 
\beq
R(\breve{g}) - R(g^*) \leq \underbrace{C \sum_{j=1}^d \| f_j - \hat{f} \|_{W^{1,1}}}_{\text{Approximation error}} + \underbrace{4d \mathfrak{R}(\CG) + 2d B_{\ell} \sqrt{\frac{\log 2/\delta}{2 n}}}_{\text{Estimation error}} + \underbrace{\kappa_1(\delta', n) + d B_{\ell} B_q \kappa_2(f - \hat{f})}_{\text{Higher-order terms}}.
\eeq
Here $\| \cdot \|_{W^{1,1}}$ is the Sobolev norm and $\mathfrak{R}(\CG)$ is the effective Rademacher complexity defined by 
\beq
\mathfrak{R}(\CG) \triangleq \frac{1}{n} \EE_{\hat{S}} \EE_{\sigma} \left[ \sup_{g\in \CG}\left| \sum_{i=1}^n \sigma_is\EE_{S_2', \cdots, S_d'}  \tilde{\ell}(\hat{s}_i, S_2', \cdots, S_d')\right| \right],
\eeq
where $\{ \sigma_i \}_{i=1}^n$ are independent sign variables, $\EE_{\hat{S}}$ is the expectation wrt $\{\hat{s}_i\}_{i=1}^n$, the dummy variables  $S_2', \cdots, S_d'$ are {\it i.i.d.} copies of $\hat{s}_1$, and $\tilde{\ell}$ is defined by 
\beq
\tilde{\ell}(s_1, \cdots, s_d) \triangleq \frac{1}{d!} \sum_{\pi} \ell(g, \hat{f}^{-1}(s_{\pi(1)}, \cdots, s_{\pi(d)})), 
\eeq
where $\pi$ denotes the permutations. $\kappa_1, \kappa_2$ are higher order terms, $B_q, B_\ell$ respectively depends on density $q$ and loss $\ell$, while $C'$ depends on ($f,q,\ell,d$). 
\end{thm}

\subsection{Speedup from shared embedding}

\citep{robinson2020strength} built some interesting theories trying to answer the following question: ``{\it Can large amounts of weakly labeled data provably help learn a better model than strong labels alone?}'' The answer is positive, assuming there is a shared embedding between the {\it weak} and {\it strong} tasks, which respectively refers auxiliary (secondary) and main tasks of interests. We summarize its main findings below and elaborate how it lends support for ECRT. 

\newcommand{\CW}{\mathcal{W}}
In the setting of weakly supervised learning, we have the triplet $(\CX, \CW, \CY)$, where $\CX$ and $\CY$ respectively denote the features and labels of interest (strong task), and $\CW$ denote weak task labels that are relevant to the prediction of $\CY$. It is assumed that there is this unknown good embedding $Z=f_0(X)$ that predicts $W$, that could be leveraged to derive a model of the form $\hat{g}(\cdot, \hat{f}): \CX \rightarrow \CY$ that improves learning. 

\begin{algorithm}[H]
   \caption{Weakly supervised learning}
   \label{alg:wsl}
\begin{algorithmic}
%\small
\STATE 1. Pretrain encoder with weak labels
%\vspace{-.7em}
$$
\hat{f} \leftarrow \text{Alg}(\CF, \PP_{XW})
$$
\vspace{-1.7em}
\STATE 2. Augment data with 
$$z_i = \hat{f}(x_i) \Rightarrow  \{ (x_i, y_i, z_i) \}\sim \hat{\PP}_{XYZ}$$
\vspace{-1.7em}
\STATE 3. Optimize the strong task
%\vspace{-.7em}
$$\hat{g} \leftarrow \text{Alg}(\CG, \hat{\PP}_{XYZ})$$
%\vspace{-1.7em}
\end{algorithmic}
\end{algorithm}

\begin{thm}[Theorem 3, \cite{robinson2020strength}]
Suppose that $\text{Rate}_m(\CF, \PP_XW) = \CO(m^{-\alpha})$ and that $\text{Alg}_n(\CG, \hat{\PP})$ is ERM. Under suitable assumptions on $(\ell, \PP, \CF)$, Algorithm \ref{alg:wsl} obtains excess risk
\beq
\CO\left(\frac{\alpha \beta \log n + \log (1/\delta)}{n} + \frac{1}{n^{\alpha \beta}} \right)
\eeq
with probability $1-\delta$, when $m = \Omega(n^{\beta})$ for $\CW$ discrete, or $m=\Omega(n^{2\beta})$ for $\CW$ continuous. 
\end{thm}

For concrete examples, in a typical learning scenario where $\text{Alg}_m(\CF, \PP_{XW}) = \CO(m^{-1/2})$, one obtains the fast rates $\CO(1/n)$ for $m=\Omega(n^2)$. 

In the context of our ECRT, we identify the learning of common causal de-mixing function $f(\zv)$ as the weak learning task, and the source space $\CS$ is the common embedding space of interest. This allows us to tape into the power of weakly supervised learning to improve the main classification task. See Figure 9 in the main text for evidence. 

\subsection{Energy-based GCL}
Mutual information (MI) is a popular metric to quantify the associations between random variables, and has been applied to a lot of areas like independent component analysis, fair learning and etc. Inspired by the FDV loss introduced in \cite{guo2021tight},
rooted from the MI between the feature $z$ and label $y$. Following the Equation (4) in the main text for $I_{\text{FDV}}$, we have a novel MI objective, pointwise mutual information (PMI) which takes the place of logistic regression objective in GCL, 
% \vspace{-0.1em}
\beq
\CL_{\text{FDV}}(f_\psi, g_\nu) \triangleq \hat{I}_{\textsc{DV}}(\{\zv_i, \yv_i\})+ \frac{\sum_j \exp[(g_\nu(\yv_j, \zv_i)-g_\nu(\yv_i, \zv_i))/\tau ]}{\sum_j \exp[(\hat{g}_\nu(\yv_j, \zv_i)-\hat{g}_\nu(\yv_i, \zv_i))/\tau ]}-1,
\label{eq:fdv-imp}
% \vspace{-1.5em}
\eeq
\vspace{-2pt}

% where $\text{sim}(\cdot, \cdot)$ is a similarity metric, which is defined as the scaled dot products and $\tau$ denotes the learnable temperature parameter. To enable the calculation of the similarity, 
% to label which also lies in the $\mathbb{R}^p$ space
% Specifically, the similarity function is defined as 
% where $\yv_i$ is the linear embedding of the original label $y_i$ into $\mathbb{R}^p$ space, $\zv_i$ is the latent representation, with dimension $m$. Denote a linear transformation matrix $\Wv^{m\times p}$. $g_{\nu}(\yv_j,\zv_i) =  \text{sim}_\nu(\yv_j,\sum_{a=1}^d[\sv]_a^\top\Wv^{m\times p})$, where $ \text{sim}_\nu(\yv,[\sv]_a^\top\Wv^{m\times p})=\frac{\yv^\top[\sv]_a^\top\Wv^{m\times p}}{\lVert\yv\rVert \lVert[\sv]_a^\top\Wv^{m\times p}\rVert}$, based on the linearity of the inner product, and the inputs are the $a$-th coordinate of $\sv=f_\psi(\zv)$ and the embedded label $\yv$. 

where $\yv_i$ is the embedding of label, $\tau$ is a learnable temperature parameter, $g_{\nu}(\yv,\zv) =  \text{sim}(\yv,\sum_{a=1}^d \gamma_\nu^a(y, [\sv]_a))$, and $ \text{sim}(\xv, \yv)=\frac{\xv^\top \yv}{\lVert \xv \rVert \lVert \yv \rVert}$, $\gamma_\nu$ and $\sv$ are the same as Equation (1) with slightly change of dimension with linear transformations.

% One notably difference between Equation (\ref{eq:fdv-imp}) and Equation (1) from the $\GCL$ is that by the linearity of the inner product, the critic loss can be computed with matrix parallelization, without looping through all the dimensions of $s$. This refinement can benefit the learning efficiency of the NICA step, as illustration in Figure~\ref{fig:corr_cmp}.

\begin{wrapfigure}{r}{0.35\textwidth}
% \begin{figure}
\vspace{-2.3em}
\begin{center}
\includegraphics[width=.30\textwidth]{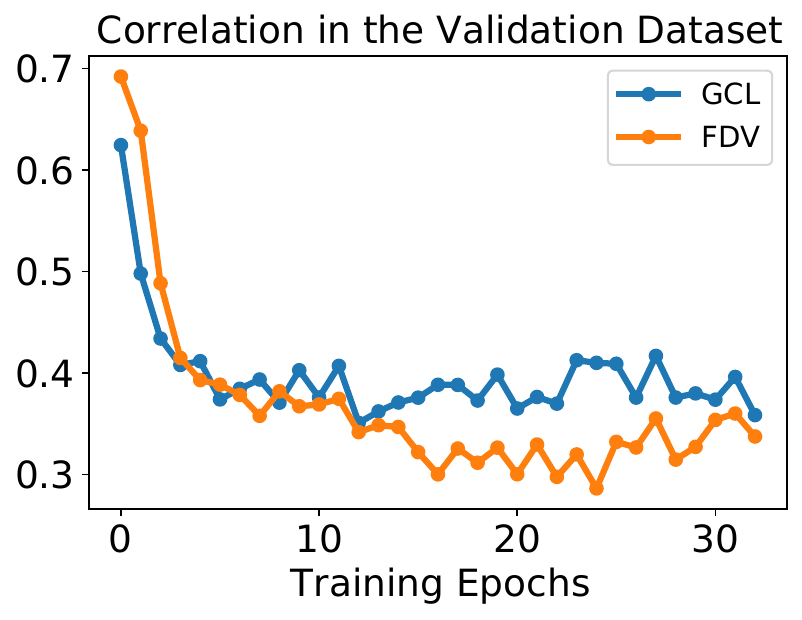}
\end{center}
\vskip -.2in
\caption{Comparison of correlation decreasing with GCL and FDV in the MNIST dataset. FDV approaches the optimality faster, which indicates better efficiency.}
\label{fig:corr_cmp}
\end{wrapfigure}

One notably difference between Equation (\ref{eq:fdv-imp}) and Equation (1) from the $\GCL$ lies in the negative sample size. In energy-based GCL, the model can treat all the examples within a minibatch as negative examples, while the original GCL contrasts with only one negative example in the minibatch by permutating labels. This refinement can benefit the learning efficiency of the NICA step, the correlation between labels decrease faster than the original GCL, as illustrated in Figure~\ref{fig:corr_cmp}.

% by the linearity of the inner product, the critic loss can be computed with matrix parallelization, without looping through all the dimensions of $s$. 

\vspace{-5pt}
\subsection{Invertible neural network}
\vspace{-3pt}
The recent interest in generative modeling has popularized the use of {\it invertible neural networks} (INN) in machine learning, with prominent examples such as normalizing flows \citep{rezende2015variational} and neural ordinary differential equations (ODEs) \citep{chen2018neural}.
Unlike standard neural networks, an INN seeks to establish a one-to-one mapping between the input and output domains, {\it i.e.}, the forward map $\sv = f_\psi(\zv)$ as well as the corresponding inverse map $\zv = f_\psi^{-1}(\sv)$.
Standard constructions of INN achieve representational flexibility by stacking simple invertible transformations. 
In practice, the efficiency of the forward or inverse passes are often trade-off depending on the application needs \citep{papamakarios2017masked}. 
% Naturally, practical trade-offs are often considered between the ease of forward and inverse passes depending on applications \citep{papamakarios2017masked}.
Here we aim for fast forward computations, thus adopt the {\it masked auto-regressive flow} (MAF) design for our INN \citep{papamakarios2017masked}, which allows efficient parallelization of the autoregressive architecture via {\it causal masking}.
% \new{allowing efficient parallel computation of an auto-regressive architecture enabled by causal masking.} 
%which is an implementation of scale-and-shift transforms that employs causal masking to enable parallel computation of an auto-regressive architecture.
% \rh{not clear, is the causal masking helping with efficiency?}

Let $\{ \zv^t \}_{t=0}^T$ be a flow of length $T$, in which $\zv^0 = \zv$, $\zv^{t+1} = F_t(\zv^{t})$, $\sv = \zv^T$, and we let $f_\psi = F_1 \circ F_2 \cdots \circ F_{T-1}$.
Specifically, the MAF is constructed as a series of {\it shift and scale} transformations $\zv^{t+1} = F_t(\zv^t) \triangleq \av_t(\zv^t) \odot \zv^t + \bv_t(\zv^t)$, where $\odot$ is the element-wise product, and $\av_t(\cdot)$ and $\bv_t(\cdot)$ are vector transformations known as scale and shift, respectively, that follow a causal autoregressive structure, {\it i.e.}, that $[\zv^{t+1}]_k$ only depends on $[\zv^t]_{<k}$.
As a direct consequence of this structure, the MAF-based INN results in a tractable Jacobian $J(\zv)\triangleq |\det(\nabla_{\zv} f_\psi)|$ of $f_\psi(\zv)$ that facilitates likelihood computations.
In fact, assuming the sources $\Sv$ have prior density $p(\sv)$, the likelihood of the features $p(\zv)$ of the MAF specification is given by \citep{papamakarios2017masked}:
%
%MAF is constructed as a series of scale-and-shift transformations in the form of $\zv^{t+1} = F_t(\zv^t) \triangleq \av_t(\zv^t) \odot \zv^t + \bv_t(\zv^t)$.
%Here $\av_t(\cdot)$ $\bv_t(\cdot)$ are vector transformations known as scale and shift terms, and they follow a specific design such that the auto-regressive property holds, {\it i.e.}, that $[\zv^{t+1}]_k$ only depends on $[\zv^t]_{<k}$.}
% scale and shift terms,
%, where $\av_t$ and $\bv_t$ are functions of $\zv^t$, respectively known as the scale and shift terms, subject to the constraint that the $i$-th output in $\av_t, \bv_t$ only depends on $[\zv^t]_{<i}$, {\it i.e.}, conforming to the causal structure of an auto-regressive flow.
%\rh{what is the output of $\av_t$? is $\av_t$ according to the description, $\av_t$ is a function of $z^t$ that returns a vector that is also called $\av_t$ with ordered entries. How do you define the order?}
%Shuffling layers are inserted to avoid order-dependent degeneracies.
%\rh{how is this done, and what degeneracies may occur?}
%For many applications of INN, generative flows in particular, also seek a tractable Jacobian $J(\zv)\triangleq |\det(\nabla_{\zv} \fv)|$ of $f(\zv)$ to enable likelihood computations.
%\rh{$\fv$ is not defined.}
%Assuming the source representation $\Sv$ has density $p(\sv)$, the data likelihood $p(\zv)$ of a scale-and-shift flow model is given by \citep{papamakarios2017masked}
%
\begin{align}\label{eq:flow}
\begin{aligned}
\log p(\zv) & = \log |\det(\nabla_{\zv} f_{\psi})| + \log p(\sv) = \sum_t \log |\av_t(\zv^t)| + \log p(\sv) \triangleq \CL_{\FLOW}(f_{\psi})
\end{aligned}
\vspace{-5pt}
\end{align}
Jointly with $\CL_{\GCL}(f_\psi, r_\nu)$ (Equation (1) in the main text), $\CL_{\FLOW}(f_{\psi})$ can be used to optimize the parameters of the de-mixing function $f_\psi(\zv)$.

\section{Regression for continuous labels}

We can further extend the applicability of the proposed ECRT to the case of regressing continuous outcomes. While in principle, the procedures described in Sec 3 can be readily applied, we advocate coarse graining wrt label $y$ similar to what has been practiced in {\it sliced inverse regression} \citep{li1991sliced}, especially when the feature dimension is high relative to the sample size. Specifically, we partition $y$ into different bins, and use feed the bin label as the conditioning variable in the GCL step. We still use the regression loss for the training of encoder and predictors.

% \section{Model Implementation}
% \subsection{Augmentation}
\section{Implementation of Augmentation}

Let $\hat{\sv}_i^{k}, i=1,\cdots, n_k$ be the estimated source representation for the $k$-th class. 

\begin{itemize}
\item Non-parametric augmentation: shuffling indices as described in the main text.
\item Parametric augmentation: estimate $\hat{\mu}_k=\text{mean}(\hat{\sv}^k), \hat{\sigma}_k=\text{std}(\hat{\sv}^k)$, then $\sv^{k,aug} \sim \CN(\hat{\mu}_k, \hat{\sigma}_k^2)$. 
\item Oracle augmentation: nonparametric augmentation with an abundance of class-conditional source space samples .
\end{itemize}
In Figure 9 from main text we compare the efficiency of parametric and nonparametric augmentation schemes under different minority sample size. In particular, we compute the MMD distance $\| \hat{\mu}_{\mathit{aug}} - \hat{\mu}_{\mathit{ref}} \|_{\kappa}$, where $\hat{\mu}_{aug} = \sum_i \kappa(\tilde{s}_i, \cdot)$ and $\hat{\mu}_{ref} = \sum_i \kappa(s_i, \cdot)$. Here $\kappa(\cdot, \cdot)$ is the Gaussian rbf kernel $\kappa(x,y)=\exp(\|x-y\|^2/2\sigma^2)$, $\|f\|_{\kappa} = \sqrt{\langle f, f \rangle}_{\kappa}$ is the RKHS norm and $\tilde{s}_i, s_i$ respectively denote augmented class-conditional samples (from few minority samples) and empirical distribution of class-conditional samples (where we use all samples from the same class that we holdout). In this example we use $2k$ samples for the ground truth and augment minority to the same size. Different kernel bandwidth $\sigma$ of $\kappa$ yields qualitatively similar results, and in the paper we report the one with $\sigma=0.5$. 

\begin{figure}
%\vspace{-3em}
\begin{center}
\includegraphics[width=.8\textwidth]{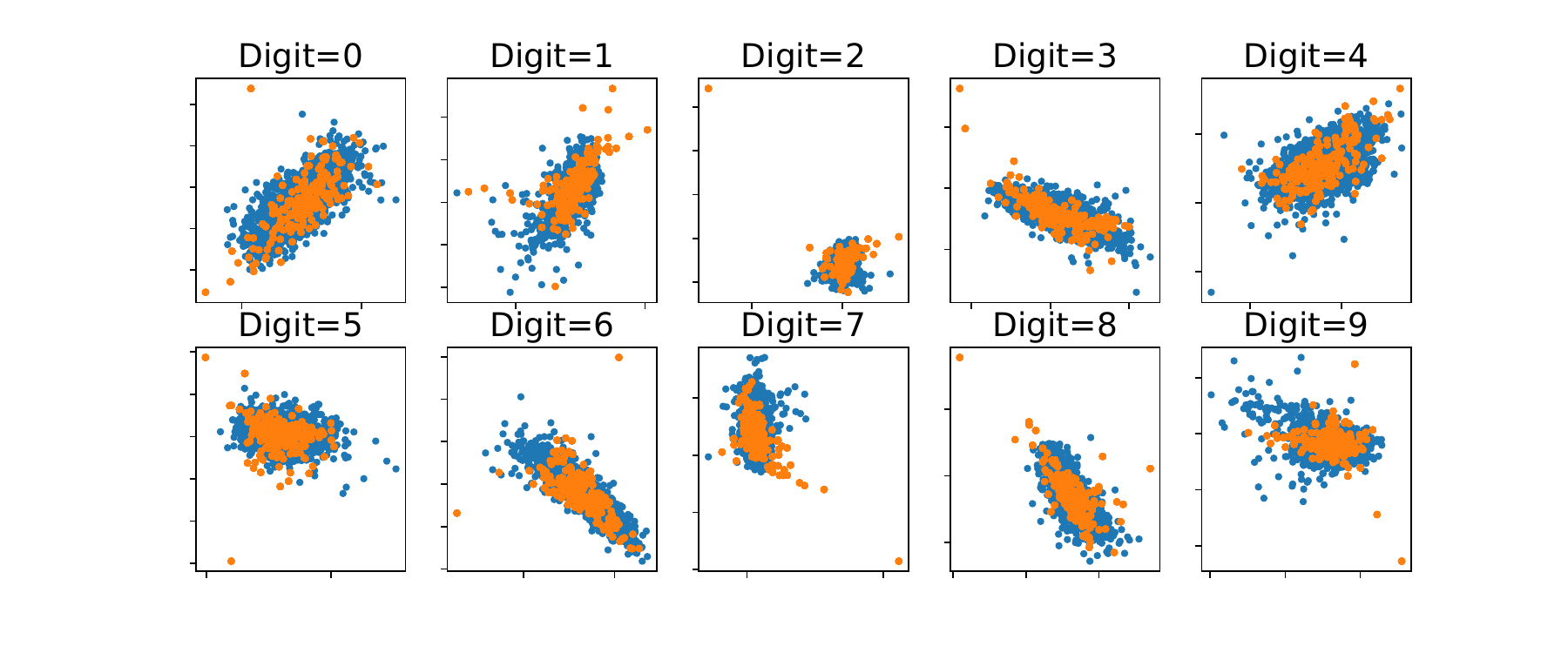}
\end{center}
\vspace{-2em}
\caption{Feature space augmentation for MNIST.}
\label{fig:mnist_z_aug}
\vspace{-1em}
\end{figure}

In Figure \ref{fig:mnist_z_aug}, we visualize the augmentation in feature space for the MNIST dataset. And we see for boundary points the discrepancy can be amplified by the neural network inversion, which partly explained the sub-optimal performance from feature space augmentation. In contrast, the source space augmentation advocated in this paper is more computationally efficient and robust.

Note that while in principle the majority classes can be similarly augmented, we choose not to refine our model with the augmented majorities. This decision is justified by the classical consideration for bias-variance trade-off: estimation errors of $f(\zv)$ is inevitable ({\it e.g.}, finite sample size, SGD, limited network capacity, {\it etc.}), and they will carry over to the augmented samples, resulting biases in the augmented estimation of our predictor. On the other hand, using augmented samples helps bring down estimation variance. For minority labels, the reduction in variance is greater than the induced bias, and consequently merits the application of ECRT to improve performance. For majority labels, this might not be the case. 

\section{Toy Model Experiment}
% we distort $\sv$ into $\tilde{\sv}$ by applying a few rotations and scaling, followed by the classical H\'enon transformation $z_1 = 1 - 1.4\, \tilde{s}_1^2 + \tilde{s}_2, z_2 = 0.3 \, \tilde{s}_1$. See Figure \ref{} (a-b) for a visualization of our toy data respectively in source and feature spaces. To render the data more realistic, we applied random affine transformation and corrupt it with random Gaussian noise, {\it i.e.}, $\xv = \Amat \zv + \epsilonv$, resulting noisy data in a high-dimensional space ($\Amat \in \BR^{32\times 2}, \epsilonv \in \BR^{32}$, draws from iid standard Gaussian). See our SM for detailed specifications. 
\subsection{Toy data demo}
We sample seven groups of two-dimensional uncorrelated Gaussian of each with size $2000$, with different means and variances as our real source representation $\sv$. Specifically, $s_i \sim N(\muv_i, \Sigmav_i), i=0, \cdots, 6$, where $\muv_0= [-0.5,-1], \muv_1 = [2,1], \muv_2 = [5,2], \muv_3 = [1,3], \muv_4=[-2,1], \muv_5=[-3.5,4], \muv_6 = [-4,-1], \Sigmav_0= [0.5,0.5], \Sigmav_1=[3,1], \Sigmav_2=[1,2], \Sigmav_3=[0.3,2], \Sigmav_4=[1, 0.2], \Sigmav_5=[1,1],\Sigmav_6= [2, 0.3]$. Then we perform classical H\'enon transformation $z_{(1)} = 1 - 1.4\, \tilde{s}_{(1)}^2 + \tilde{s}_{(2)}, z_{(2)} = 0.3, \tilde{s}_{(1)}$ to generate the data in feature space.

\subsection{Extreme-classification Toy Data}
We sample 1000 groups of two-dimensional uncorrelated Gaussian with mean ranges uniformly sampled from range $(-4,4)$ and standard deviation fixed to $0.1$. Validation dataset is fixed with 20 samples per-class, and the sampler per class in training dataset varies with $5, 10, 20, 50, 75, 100, 150$. The summary for extreme-classification is presented in Table~\ref{tab:extremeValid}. 

Note that when the total number of categories is 10 with 20 samples per class, ERM has top-1 accuracy as high as 0.914, and the performance drops when the number of categories increasing. With 500 categories, the accuracy decreases to 0.005, and in the scenario where we presented in the main text with 1000 categories, ERM performs no better than random guessing.

\begin{table}[ht]
\centering
\caption{Validation Results for Extreme classification}
\label{tab:extremeValid}
\begin{sc}
\resizebox{.9\textwidth}{!}{
\begin{tabular}{c|ccc|ccc|ccc}
\hline
Metric & \multicolumn{3}{c}{NLL}     & \multicolumn{3}{c}{Top 1}   & \multicolumn{3}{c}{Top5}    \\ \hline
Sample-size      & ERM    & ECRT W/O Aug & ECRT    & ERM    & ECRT W/O Aug & ECRT    & ERM    & ECRT W/O Aug & ECRT    \\ \midrule
5          & 6.9246 & 4.3884    & {\bf 3.7098} & 0.0015 & 0.0450    & {\bf 0.0760} & 0.0053 & 0.2275    & {\bf 0.3220} \\
10         & 6.9242 & 4.0536    & {\bf 3.4752} & 0.0018 & 0.0638    & {\bf 0.0856} & 0.0060 & 0.2666    & {\bf 0.3549} \\
20         & 6.9185 & 3.5405    & {\bf 3.3657} & 0.0016 & 0.0741    & {\bf 0.0877} & 0.0070 & 0.3126    & {\bf 0.3675} \\
50         & 6.9240 & 3.3794    & {\bf 3.2957} & 0.0010 & 0.0834    & {\bf 0.0934} & 0.0045 & 0.3493    & {\bf 0.3835} \\
75         & 6.9190 & 3.3499    & {\bf 3.2594} & 0.0020 & 0.0846    & {\bf 0.0939} & 0.0060 & 0.3571    & {\bf 0.3950} \\
100        & 6.9182 & 3.3394    & {\bf 3.2675} & 0.0020 & 0.0834    & {\bf 0.0934} & 0.0053 & 0.3580    & {\bf 0.3918} \\
150        & 6.9174 & 3.2731    & {\bf 3.2597} & 0.0011 & 0.0906    & {\bf 0.0984} & 0.0059 & 0.3777    & {\bf 0.4034} \\ \hline
\end{tabular}
}
\end{sc}
\end{table}

\section{Real-world Data Experiments}

\subsection{Image Data Benchmarks}

We summarized the image datasets in Table \ref{tab:data_size}  and the network architectures used for respective datasets in Tables \ref{tab:mnist_net}, \ref{tab:cifar_net}, \ref{tab:inat_net} and \ref{tab:tiny_net}. The hyperparameters we used in these experiments are presented in \ref{tab:hyper}. The results reported here are from our regularized non-parametric ECRT implementation, parametric ECRT implementation show a similar trend, with slightly improved performance (results now shown). We use $2$ latent dims for MNIST and $32$ latent dims for CIFAR100, iNaturalist and Tiny-Imagenet.

\begin{figure}[ht]
%\vspace{-3em}
\begin{minipage}{.5\linewidth}
\begin{center}
\vspace{-2em}
\includegraphics[width=.7\textwidth]{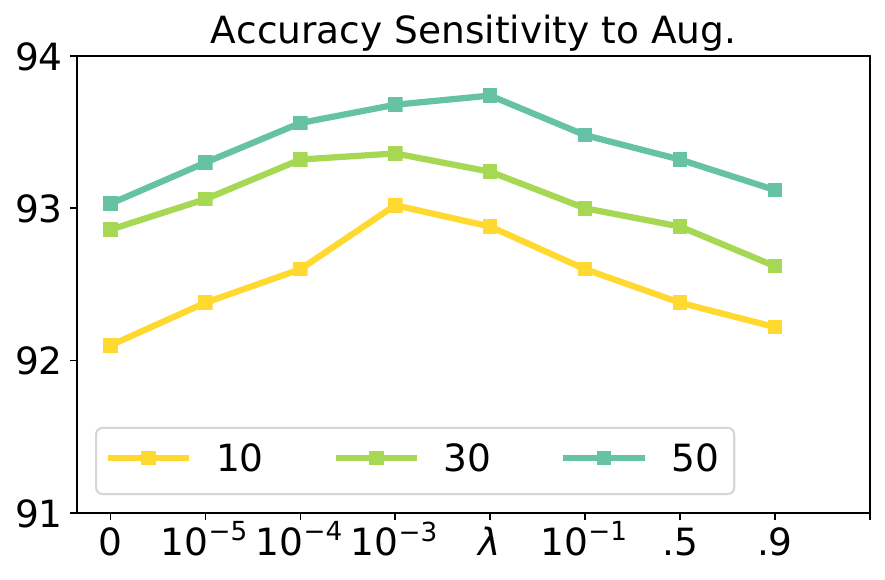}
\end{center}
\vspace{-10pt}
\caption{Sensitivity analysis (overall top 1 accuracy) of augmentation strength $\lambda$. Complementing Figure 11 in main text.}
\label{fig:acc_aug}
\end{minipage}
\begin{minipage}{.5\linewidth}
\begin{center}
\vspace{10pt}
\includegraphics[width=.7\textwidth]{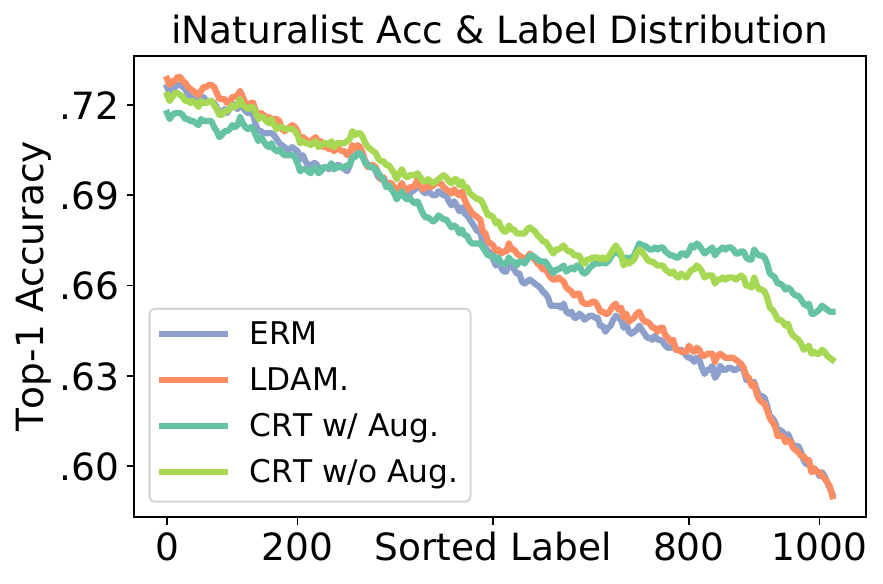}
\end{center}
\vspace{-10pt}
\caption{Class-conditional Top-1 accuracy curve for iNat2019. Complementing Figure 1 (F1 score). Note that ERM and LDAM show better accuracy for sample-rich majorities, but worse F1 scores. This evidences majority bias, that a predictor has a low specificity for data rich classes.}
\label{fig:acc_aug}

\end{minipage}
\vspace{-1em}
\end{figure}

{\bf Preprocessing}
For iNaturalist dataset, we used pretrained Inception V3 to extract the features with dim=$2048$. For Tiny-imagenet we finetuned the Resnet 18 to extract the features with dim = $512$.

{\bf Baselines}
We used the ERM, LDAM \footnote{\url{https://github.com/kaidic/LDAM-DRW}}, Focal \footnote{\url{https://github.com/artemmavrin/focal-loss}}, IW \footnote{\url{https://github.com/idiap/importance-sampling}}, and VAT \footnote{ \url{https://github.com/lyakaap/VAT-pytorch}} baseline implementations. For GAN we adopted the CGAN model with architecture shown in \ref{tab:GAN} and noise dimension listed in \ref{tab:hyper_gan} for each dataset. We compared all the models with their own best performance after early stopping.

{\bf Discrepancy of baseline performance.} We noticed that our implementation of baseline models, especially for the ERM baseline, yields results better than what's reported in literature (LDAM in particular). Specifically, our results look better. After carefully compared our implementation to the LDAM codebase, we see that the discrepancy comes from the choice of optimizer. The use of vanilla SGD optimizer, as practiced in LDAM, results in degraded performance of baseline, and consequently a larger performance gap compared to strong solutions.

{\bf Majority bias.} In Figure \ref{fig:acc_aug}, we give the top-1 accuracy wrt different minority size on the iNaturalist dataset. This figure is complementary to the F1-label frequency plot given in Figure 14 from the main text. While the improvement at the tail part are strong under both metrics, we see clear evidence of majority bias in the Top-1 accuracy plot. ERM and LDAM show better performance in accuracy for the sample abundant majority regime, but severe performance drop in the sample deficient minority regime. This is because ERM and alike finds it more rewarding to favor the majorities during inference, which gives better sensitivity but much worse specificity for the majority samples, and consistently hurting the performance for minorities. 

\subsection{Language Data Benchmark}

{\bf Dataset and preprocessing.} In this experiment, have used the \texttt{arXiv} dataset hosted on \texttt{Kaggle} \footnote{\url{https://www.kaggle.com/Cornell-University/arxiv}}. We use the pretrained BERT model from the \texttt{transformers} package \footnote{\url{https://github.com/huggingface/transformers}} to extract sentence features. Specifically, we applied the \texttt{SciBERT} model  (\texttt{allenai/scibert\_scivocab\_uncased}) \citep{beltagy2019scibert}, and used the BERT default $768$-dimensional sentence embedding for each abstract.  The training set includes 160k data, where class labels with more than 5k samples are identified as majority classes, with the rest assigned to minority label classes. All label classes with less than 20 samples have been excluded from our analysis. This gives us a total of 14 majority classes and 138 minority classes.

{\bf Setup.} Different from the image benchmarks, the \texttt{arXiv} data is a multi-label prediction task. Each abstract is associated with at least one, possibly multiple labels, and we make binary classifications for each label class.  In the training of GCL model, we allow samples with multiple labels to be reused by different classes, as each constructs a valid source IC distribution under our hypothesis. Only standard ECRT is considered in this experiment. We set the source space dimension to $64$ and use the network architecture described in Table \ref{tab:arxiv_net}. 

{\bf Evaluation.} The accuracy reported for this experiment is defined as follows: say a sample is associated with $k$-labels, then we compared the top-$k$ predicted labels to the true labels, and report the averaged accuracy for this sample. Like previous experiments, we target a balanced evaluation set. However, getting a perfectly balanced evaluation set is impossible here, as samples are associated with multiple labels. We extracted a nearly-balanced evaluation set including 847 samples, where each label has 10 to 50 counts. Most of classes have 10-15 samples in our nearly-balanced evaluation set.

\begin{table}[ht]
\caption{Summary of datasets \label{tab:data_size} }
\setlength{\tabcolsep}{15pt}
\begin{center}
\begin{small}
\begin{sc}
\resizebox{.9\textwidth}{!}{
\begin{tabular}{lcccccccccc}
\toprule
&Name  &Dim & Train (Majority) & Train (Minority) & Validation \\
\midrule
& MNIST & ($28\times28$) & $6000$ $\times (1$ or $5$) (cls) & $1200$ $\times (1$ or $5$) (cls) & $1000 \times 10$ (cls) \\

& Cifar & ($32\times32\times3$) & $500$ $\times 50$ (cls) & $500$ $\times 50$ (cls) & $100 \times 100$ (cls) \\

& INat & (None $\times$ None$ \times3$) & $(\geq 120)$ $\times 725$ (cls) & $(<120)$ $\times 285$ (cls) & $3 \times 1010$ (cls) \\

& Tiny & ($64 \times 64 \times3$) & $450$ $\times 100$ (cls) & $45$ $\times 100$ (cls) & $50 \times 200$ (cls) \\

& Arxiv & ($None$) & $(>5000)$ $\times 14$ (cls) & $(<5000)$ $\times 138$ (cls) & $12 \times 152$ (cls) \\
\bottomrule
\end{tabular}
}
\end{sc}
\end{small}
\end{center}
\end{table}

\begin{table}[!t]
    % \caption{Global caption}
\begin{minipage}{.5\linewidth}
\caption{MNIST experiment network architecture.\label{tab:mnist_net}}
\setlength{\tabcolsep}{25pt}
\centering
\begin{sc}
\begin{small}
\resizebox{.9\textwidth}{!}{\begin{tabular}{clcc}
\toprule
\multicolumn{1}{c}{Network} & \multicolumn{3}{c}{Architecture} \\
\midrule
Encoder & fc(unit=32)+ReLU \\
&+ fc(unit=32)+ReLU\\
&+ fc(unit=2)\\
[5pt]
Decoder & fc(unit=32)+ReLU \\
&+ fc(unit=32)+ReLU\\
&+ fc(unit=10)\\
[5pt]
FLOW & MAF($n_{blocks} = 4,$\\
&$hidden_{size}=128,$\\
& $n_{hidden}=2$).\\
\bottomrule
\end{tabular}}
\end{small}
\end{sc}
\end{minipage}%
    \begin{minipage}{.5\linewidth}
      \centering
\caption{Cifar100 experiment network architecture. \label{tab:cifar_net} }
\setlength{\tabcolsep}{25pt}
\begin{sc}
\begin{small}
\resizebox{.9\textwidth}{!}{
\begin{tabular}{clcc}
\toprule
\multicolumn{1}{c}{Network} & \multicolumn{3}{c}{Architecture} \\
\midrule
Encoder & Resnet18 \footnote{Resnet 18 without last layer} \\
&+ fc(unit=32)\\
[5pt]
Decoder & fc(unit=256)+ReLU\\
& fc(unit=100)\\
[5pt]
FLOW & MAF($n_{blocks} = 4,$\\
&$hidden_{size}=128,$\\
& $n_{hidden}=2$).\\
\bottomrule
\end{tabular}}
\end{small}
\end{sc}
\end{minipage} 
\end{table}

\begin{table}[!t]
\begin{minipage}{.5\linewidth}
\centering
\begin{small}
\begin{sc}
\caption{iNaturalist experiment network architecture. \label{tab:inat_net} }
\setlength{\tabcolsep}{20pt}
\resizebox{.8\textwidth}{!}{
\begin{tabular}{clcc}
\toprule
\multicolumn{1}{c}{Network} & \multicolumn{3}{c}{Architecture} \\
\midrule
Pretrain & inception(v3) \\
[5pt]
Encoder & fc(unit=1024)+ReLU\\
&+Dropout(0.1) \\
&+ fc(unit=512)+ReLU\\
&+Dropout(0.1)\\
&+ fc(unit=32)\\
[5pt]
Decoder & fc(unit=32)+ReLU\\
&+Dropout(0.1) \\
&+ fc(unit=512)+ReLU\\
&+Dropout(0.1)\\
&+ fc(unit=1010)\\
[5pt]
FLOW & MAF($n_{blocks} = 4,$\\
&$hidden_{size}=128,$\\
& $n_{hidden}=2$).\\
\bottomrule
\end{tabular}
}
\end{sc}
\end{small}
\end{minipage}
\hspace{2pt}
\begin{minipage}{.5\linewidth}
\centering
\begin{small}
\begin{sc}
\caption{Tiny ImageNet experiment network architecture. \label{tab:tiny_net} }
\setlength{\tabcolsep}{20pt}
\resizebox{.8\textwidth}{!}{
\begin{tabular}{clcc}
\toprule
\multicolumn{1}{c}{Network} & \multicolumn{3}{c}{Architecture} \\
\midrule
Pretrain & ResNet18 \\
[5pt]
Encoder & fc(unit=1024)+ReLU\\
&+Dropout(0.1) \\
&+ fc(unit=512)+ReLU\\
&+Dropout(0.1)\\
&+ fc(unit=32)\\
[5pt]
Decoder & fc(unit=512)+ReLU\\
&+Dropout(0.1) \\
&+ fc(unit=200)\\
[5pt]
FLOW & MAF($n_{blocks} = 4,$\\
&$hidden_{size}=128,$\\
& $n_{hidden}=2$).\\
\bottomrule
\end{tabular}}
\end{sc}
\end{small}
\end{minipage} 
\end{table}

\begin{table}[!t]
\begin{minipage}{.5\linewidth}
\centering
\begin{small}
\begin{sc}
\caption{Arxiv experiment network architecture. \label{tab:arxiv_net} }
\setlength{\tabcolsep}{20pt}
\resizebox{.8\textwidth}{!}{
\begin{tabular}{clcc}
\toprule
\multicolumn{1}{c}{Network} & \multicolumn{3}{c}{Architecture} \\
\midrule
Pretrain &  BERT\\
[5pt]
Encoder & fc(unit=1024)+ReLU\\
&+Dropout(0.1) \\
&+ fc(unit=512)+ReLU\\
&+Dropout(0.1)\\
&+ fc(unit=64)\\
[5pt]
Decoder & fc(unit=64)+ReLU\\
&+Dropout(0.1) \\
&+ fc(unit=512)+ReLU\\
&+Dropout(0.1)\\
&+ fc(unit=152)\\
[5pt]
FLOW & MAF($n_{blocks} = 4,$\\
&$hidden_{size}=128,$\\
& $n_{hidden}=2$).\\
\bottomrule
\end{tabular}
}
\end{sc}
\end{small}
\end{minipage} 
\begin{minipage}{.45\linewidth}
\centering
\label{tab:MNISTresult}
\caption{MNIST results with different numbers of minority categories}
\resizebox{\linewidth}{!}{
\begin{sc}
\begin{tabular}{@{}ccccc@{}}
\toprule
\# minority label & \multicolumn{2}{c}{1} & \multicolumn{2}{c}{5} \\ \midrule
                  & NLL       & Top 1     & NLL       & Top 1     \\\midrule
ERM               & 0.342     & 0.933     & 0.390     & 0.904     \\
LDAM              & 1.6737    & 0.9609    & 1.50      & 0.940     \\
ECRT               & {\bf 0.186}     & {\bf 0.972}     & {\bf 0.257}     & {\bf 0.950}     \\ \bottomrule
\end{tabular}
\end{sc}
}
\end{minipage}
\end{table}

% \begin{table}[!t]
% \begin{minipage}{.45\linewidth}
% \centering
% \label{tab:MNISTresult}
% \caption{MNIST results with different numbers of minority categories}
% \resizebox{\linewidth}{!}{
% \begin{sc}
% \begin{tabular}{@{}ccccc@{}}
% \toprule
% \# minority label & \multicolumn{2}{c}{1} & \multicolumn{2}{c}{5} \\ \midrule
%                   & NLL       & Top 1     & NLL       & Top 1     \\\midrule
% ERM               & 0.342     & 0.933     & 0.390     & 0.904     \\
% LDAM              & 1.6737    & 0.9609    & 1.50      & 0.940     \\
% ECRT               & {\bf 0.186}     & {\bf 0.972}     & {\bf 0.257}     & {\bf 0.950}     \\ \bottomrule
% \end{tabular}
% \end{sc}
% }
% \end{minipage}
% \end{table}

\begin{table}[!t]
\begin{minipage}{.5\linewidth}
\centering
% \begin{small}
\begin{sc}
\caption{GAN network architecture. \label{tab:GAN} }
\setlength{\tabcolsep}{20pt}
\resizebox{1.\textwidth}{!}{
\begin{tabular}{cc}
\toprule
\multicolumn{1}{c}{Network} & \multicolumn{1}{c}{Architecture} \\
\midrule
Generator & FC(unit=256) +LeakyRelu \\
&+ FC(unit=256)+Leakyrelu\\
&+FC(unit=latent$_{dim}$)\\
[5pt]
Discriminator &
FC(unit=256) +LeakyRelu \\
&+dropout(0.1)\\
&+ FC(unit=256)+LeakyRelu\\
&+dropout(0.1)\\
&+FC(unit=1)+sigmoid\\
\bottomrule
\end{tabular}
}
\end{sc}
% \end{small}
\end{minipage}
\begin{minipage}{.5\linewidth}
\centering
\begin{small}
\begin{sc}
\caption{Hyperparameter of datasets (ECRT)  \label{tab:hyper} }
\setlength{\tabcolsep}{20pt}
\resizebox{.8\textwidth}{!}{
\begin{tabular}{lcc}
\toprule
Name  & Reg Weight & Aug Strength\\
\midrule
MNIST & 1e-2 & 1e-3 \\
Cifar & 1e-2 & 1e-3\\
INat & 5e-3 & 1e-3\\
Tiny & 1e-3 & 1e-3\\
Arxiv & 1e-2 & 1e-2\\
\bottomrule
\end{tabular}}
\end{sc}
\vspace{5pt}
\begin{sc}
\caption{Hyperparameter of datasets (GAN)\label{tab:hyper_gan} }
\setlength{\tabcolsep}{20pt}
\resizebox{.6\textwidth}{!}{
\begin{tabular}{cc}
\toprule
Name  & noise dim \\
\midrule
MNIST & 32 \\
Cifar & 64 \\
INat & 128\\
Tiny & 128\\
arxiv & 64\\
\bottomrule
\end{tabular}}
\end{sc}
\end{small}
\end{minipage} 
\end{table}

% \begin{figure}
% %\vspace{-3em}
% \begin{center}
% \includegraphics[width=.45\textwidth]{figures/results/acc-inat-cmp.pdf}
% \end{center}
% \vskip -.1in
% \caption{Class-conditional Top-1 accuracy curve for iNat2019. Complementing Figure 1 (F1 score). Note that ERM and LDAM show better accuracy for sample-rich majorities, but worse F1 scores. This evidences majority bias, that a predictor has a low specificity for data rich classes.}
% \label{fig:acc_aug}
% \vspace{-1em}
% \end{figure}

%  \subsection{Regression for continuous labels}
%  We further extend the applicability of the proposed ECRT to the case of regressing continuous outcomes. While in principle, the procedures described in Sec \ref{sec:ECRT} and above can be readily applied, we advocate coarse graining wrt label $y$ similar to what has been practiced in {\it sliced inverse regression} \citep{li1991sliced}, especially when the feature dimension is high relative to the sample size. Specifically, we partition $y$ into different bins, and use feed the bin label as the conditioning variable in the GCL step. We still use the regression loss for the training of encoder and predictors. 

% \section{Additional Results}
% \subsection{Extreme Classification}

\clearpage
{\small
\bibliographystyle{plain}
\bibliography{crt}
%\bibliography{egbib}
}

%% file: math_cmds.tex
\newcommand{\av}{{\boldsymbol a}}

\newcommand{\fv}{{\boldsymbol f}}

\newcommand{\ov}{{\boldsymbol o}}

\newcommand{\sv}{{\boldsymbol s}}

\newcommand{\xv}{{\boldsymbol x}}
\newcommand{\yv}{{\boldsymbol y}}
\newcommand{\Xv}{{\boldsymbol X}}

\newcommand{\zv}{{\boldsymbol z}}

\newcommand{\kappav}{{\boldsymbol \kappa}}

\newcommand{\EE}{\mathbb{E}}

\newcommand{\bs}[1]{\boldsymbol{#1}}
\newcommand{\BR}{\mathbb{R}}

\newcommand{\CX}{\mathcal{X}}

\newcommand{\CZ}{\mathcal{Z}}

\newcommand{\CS}{\mathcal{S}}

\DeclareMathOperator*{\argmin}{arg\,min}

\newcommand{\CL}{\mathcal{L}}

\theoremstyle{plain}
\newtheorem{thm}{Theorem}[section] % reset theorem numbering for each chapter
\theoremstyle{definition}
\newtheorem{assumption}[thm]{Assumption}

\newcommand{\beq}{\begin{equation}}
\newcommand{\eeq}{\end{equation}}
\newcommand{\beqs}{\begin{eqnarray}}
\newcommand{\eeqs}{\end{eqnarray}}
\newcommand{\barr}{\begin{array}}
\newcommand{\earr}{\end{array}}